\documentclass[10pt,twocolumn,letterpaper]{article}

\pdfoutput=1

\usepackage{cvpr}
\usepackage{times}
\usepackage{epsfig}
\usepackage{graphicx}
\usepackage{amsmath}
\usepackage{amssymb}
\usepackage{rotating}
\usepackage{url}
\usepackage{textcomp}
\usepackage{tabularx}
\usepackage{subfigure}

\usepackage[pagebackref=true,breaklinks=true,letterpaper=true,colorlinks,bookmarks=false]{hyperref}

\cvprfinalcopy % *** Uncomment this line for the final submission

\def\cvprPaperID{153} % *** Enter the CVPR Paper ID here

\graphicspath{{./figures/}}

\newcommand{\ignorethis}[1]{}

\newcommand{\argmin}{\operatornamewithlimits{argmin}}

\newcommand\eq[1]{Eq.~(\ref{#1})}
\newcommand\sect[1]{Section~\ref{#1}}

\newcommand\fig[1]{Figure~\ref{#1}}
\newcommand\tabfig[1]{Table~\ref{#1}}

\newcommand{\transp}[1]{{#1}^\mathsf{T}}

\newcommand{\shape}{\mathbf{s}}
\newcommand{\littleFeat}{d}
\newcommand{\featDesc}{\mathbf{d}}
\newcommand{\shapeRegressor}{\mathbf{R}}

\newcommand{\imgIndex}{i}

\newcommand{\image}{I}
\newcommand{\shapesMat}{\mathbf{S}}
\newcommand{\shapesMatSymb}{\mathbf{s}}
\newcommand{\shapeBasesMat}{\mathbf{B}}

\newcommand{\shapeParam}{p}
\newcommand{\shapeParamVec}{\mathbf{\shapeParam}}
\newcommand{\numBases}{m}
\newcommand{\pnt}{s}
\newcommand{\pntIndex}{i}
\newcommand{\visVec}{\mathbf{v}}
\newcommand{\visMat}{\mathbf{V}}
\newcommand{\combParamVec}{\mathbf{q}}
\newcommand{\combBasesMat}{\mathbf{C}}
\newcommand{\cascadeIndex}{t}
\newcommand{\numCascadeLevels}{T}
\newcommand{\bcrNodeIndex}{k}
\newcommand{\nodeMems}{N}
\newcommand{\numTrees}{M}
\newcommand{\treeIndex}{m}

\newcommand{\numRandFeats}{F}

\ifcvprfinal\fi
\begin{document}

%%%%%%%%% TITLE
\title{\vspace{-12pt}Efficient Branching Cascaded Regression \\ for Face Alignment under Significant Head Rotation}

\author{Brandon M. Smith \ \ \ \ \ \ \ \ Charles R. Dyer\\
{\tt\small bmsmith@cs.wisc.edu \ \ dyer@cs.wisc.edu} \\
University of Wisconsin--Madison\\
}

\maketitle
%\thispagestyle{empty}

%%%%%%%%% ABSTRACT
\begin{abstract}
Despite much interest in face alignment in recent years, the large majority of work has focused on near-frontal faces.
Algorithms typically break down on profile faces, or are too slow for real-time applications. 
In this work we propose an efficient approach to face alignment that can handle $180$ degrees of head rotation in a unified way (\eg, without resorting to view-based models) using 2D training data. 
The foundation of our approach is cascaded shape regression (CSR), which has emerged recently as the leading strategy. 
We propose a generalization of conventional CSRs that we call branching cascaded regression (BCR).
Conventional CSRs are single-track; that is, they progress from one cascade level to the next in a straight line, with each regressor attempting to fit the entire dataset.
We instead split the regression problem into two or more simpler ones after each cascade level. 
Intuitively, each regressor can then operate on a simpler objective function (\ie, with fewer conflicting gradient directions). 
Within the BCR framework, we model and infer pose-related landmark visibility and face shape simultaneously using Structured Point Distribution Models (SPDMs). 
We propose to learn task-specific feature mapping functions that are adaptive to landmark visibility, and that use SPDM parameters as regression targets instead of 2D landmark coordinates. 
Additionally, we introduce a new in-the-wild dataset of profile faces to validate our approach. 
%Using our approach, we demonstrate excellent results on challenging datasets. 
\end{abstract}

%%%%%%%%% BODY TEXT

\section{Introduction}

There has been significant research interest in face alignment in recent years, which has resulted in impressive advances in speed and robustness on challenging in-the-wild faces. % (\eg, \cite{Kazemi2014,Lee2015,Ren2014,Xiong2015,Tzimiropoulos2015,Wu2015,Yu2014,Zhu2015}).
However, relatively little work (with the exception of a few, \eg \cite{Wu2015}) has been done to handle the full range of head poses encountered in the real world. 
The large majority of face alignment algorithms focus on near fronto-parallel faces, and break down on profile faces.
Handling the full range of head poses is challenging because:
\begin{itemize}
\item The 2D shapes of profile faces and frontal faces are significantly different. 
\item Many landmarks become self-occluded on profile faces. 
\item Algorithms must contend with greater background noise on profile faces, as most conventional landmarks (\eg, nose tip, lip corners) are not surrounded by the face. 
\end{itemize}
A straightforward approach to large head pose variation is to employ view-based models and choose the result that achieves the best match \cite{Cootes2000,Zhu2012}. 
However, because head pose is continuous, problems arise at the transition points between models (\eg, $45$ degrees), and employing many view-specific models is inefficient. 
Nonlinear statistical models (\eg, kernel methods \cite{Romdhani1999} or mixture models \cite{Cootes1997,Kanaujia2006,Zhou2005}) have been proposed to deal with large head pose variation. 
However, these methods are too slow for real-time applications. 
Another strategy is to employ a full 3D model of the face \cite{Baltrusaitis2012,Cao2013,Cao2014,Gu2006,Matthews2007}, but publicly available training data are limited, and algorithms still tend to focus on near fronto-parallel faces \cite{Cao2013,Cao2014}. 

\begin{figure}
\begin{center}
\begin{small}
\begin{tabular}{c@{\hspace{6pt}}c@{\hspace{6pt}}c}
\includegraphics[width=0.9in]{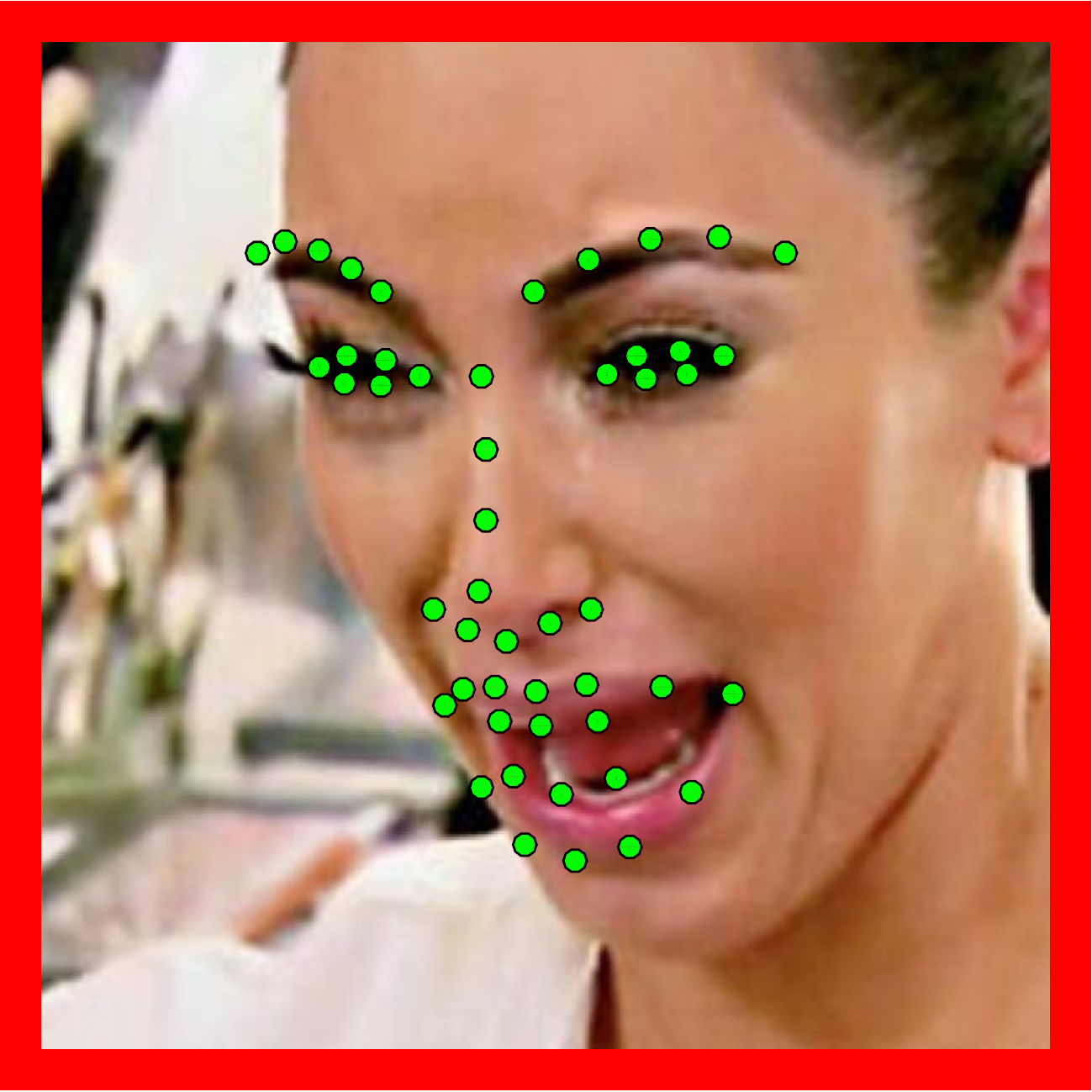}
&
\includegraphics[width=0.9in]{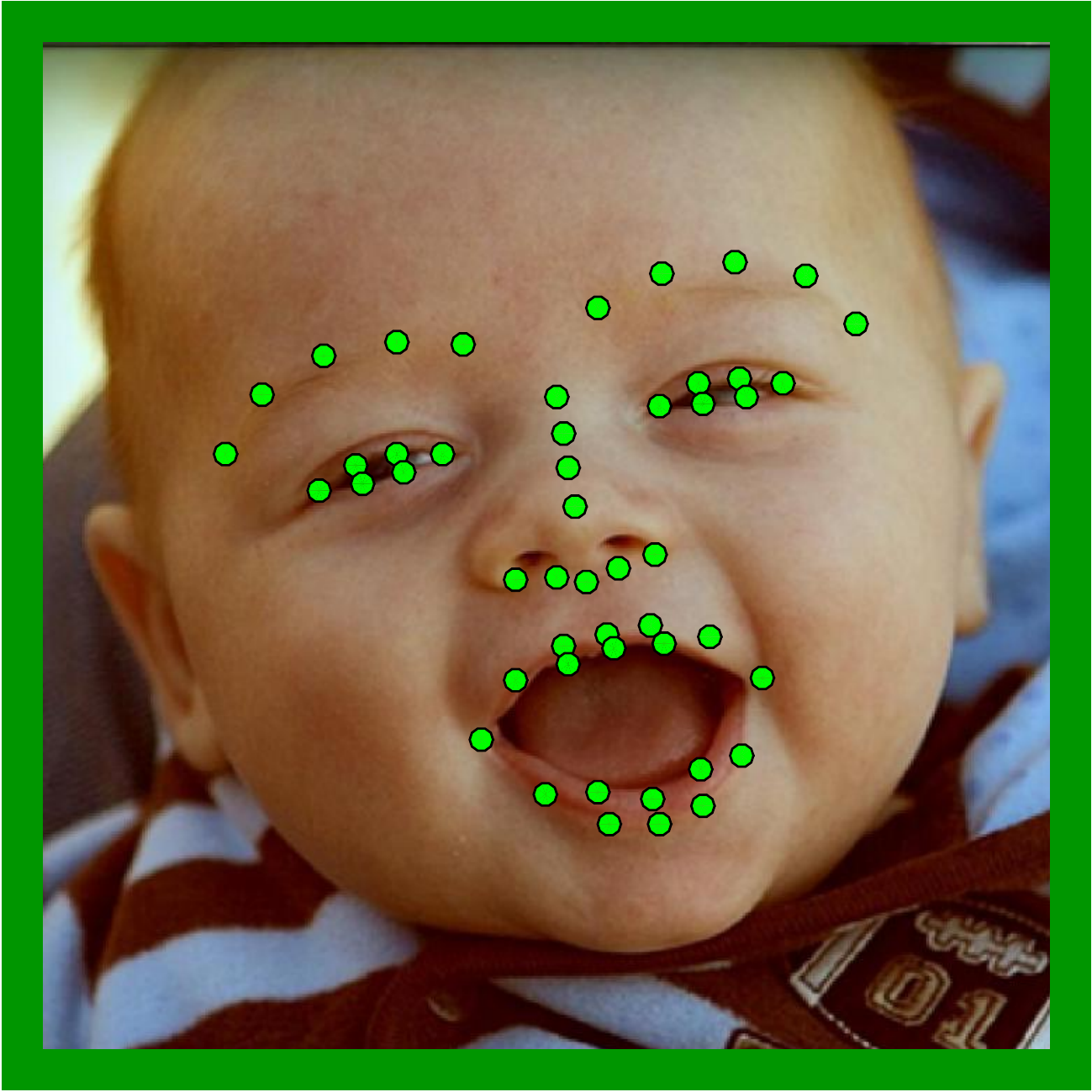}
&
\includegraphics[width=0.9in]{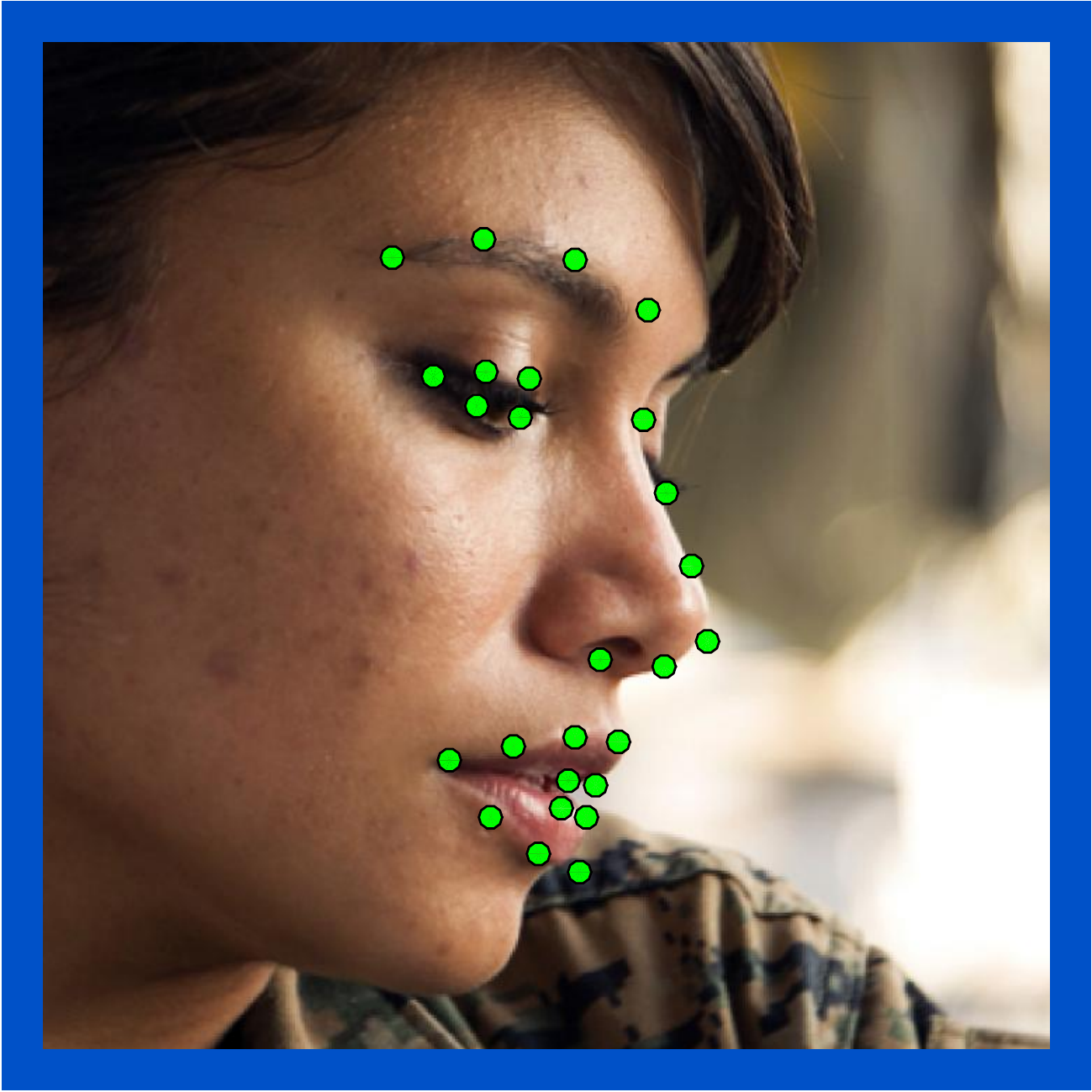}
\\
\includegraphics[width=0.95in]{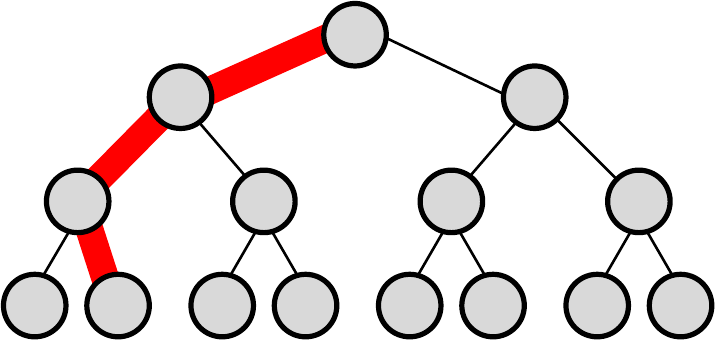}
&
\includegraphics[width=0.95in]{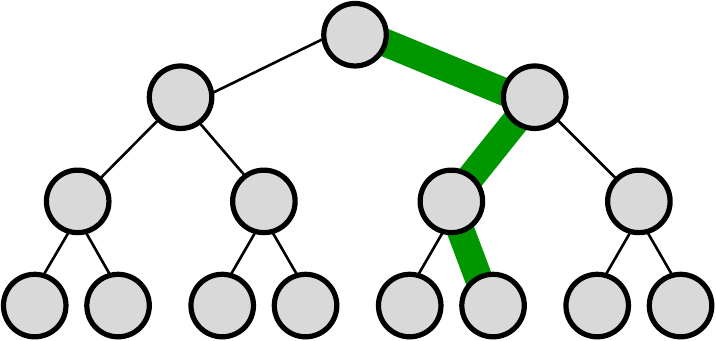}
&
\includegraphics[width=0.95in]{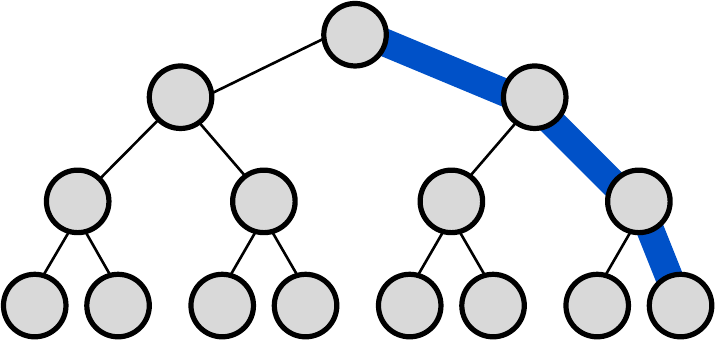}
\end{tabular}
\end{small}
\end{center}
\caption{
Our branching cascaded regression (BCR) algorithm can handle faces with a wide variety of poses, including full profile. 
} 
\label{fig-teaser} %\vspace{-8pt}
\end{figure}

We propose an approach to face alignment that can handle $180$ degrees of head rotation angle in a unified way (\eg, without resorting to view-based models) using 2D training data and linear shape models. 
The foundation of our approach is cascaded shape regression, which has emerged recently as the leading strategy \cite{BurgosArtizzu2013,Cao2012,Dollar2010,Kazemi2014,Ren2014,Tzimiropoulos2015,Wu2015,Xiong2013,Xiong2015,Yu2014}.
Cascaded shape regressors (CSRs) are computationally efficient and the core idea is elegant.
CSRs are purely discriminative, which means they can capitalize on large and diverse training datasets to perform robust face alignment in the wild. 
%Starting from a rough initial shape $\shape^{0}$, CSRs progressively update the shape estimate. 
%The input of each regression stage $t$ is the current shape estimate $\shape^{t}$ and a feature descriptor $\featDesc(\image,\shape^{t})$ that captures the local appearance in image $\image$ relative to $\shape^{t}$. The output is an updated shape: 
%\begin{eqnarray}
%\Delta \shape^{t} &=& \shapeRegressor^{t}\featDesc(\image, \shape^{t}) \\
%\shape^{t+t} &=& \shape^{t} + \Delta \shape^{t}
%\end{eqnarray}
%$\shapeRegressor^{t}$ is a regressor trained offline to minimize the following:
%\begin{equation}
%\argmin_{\hat{\shapeRegressor}^{t}} \sum_{\imgIndex}^{\numImages} \| \Delta \shape_{\imgIndex}^{*} - \shapeRegressor^{t}\featDesc(\shape_{\imgIndex}^{t}) \|^{2}, 
%\end{equation}
%where $\Delta \shape_{\imgIndex}^{*} = \shape_{\imgIndex}^{*} - \shape_{\imgIndex}^{t}$ is the ideal shape update for face $\imgIndex$ and the summation is over $\numImages$ training faces. 
However, 
we observe two key limitations with the conventional formulation of CSRs:
\begin{description}
\item[Limitation 1:] They assume that all landmarks are visible, which is problematic for profile faces. 
\item[Limitation 2:] They are local algorithms, and they are likely to average conflicting gradient directions \cite{Xiong2015}, which is especially problematic when they are applied to face collections with large head pose variation. 
\end{description}

\begin{figure}
\begin{center}
\begin{small}
\begin{tabular}{r@{\hspace{4pt}}c}
%& Left profile & & Right profile \\
\begin{sideways}\hspace{30pt}Fraction of test faces\end{sideways}
&
\includegraphics[width=2.7in]{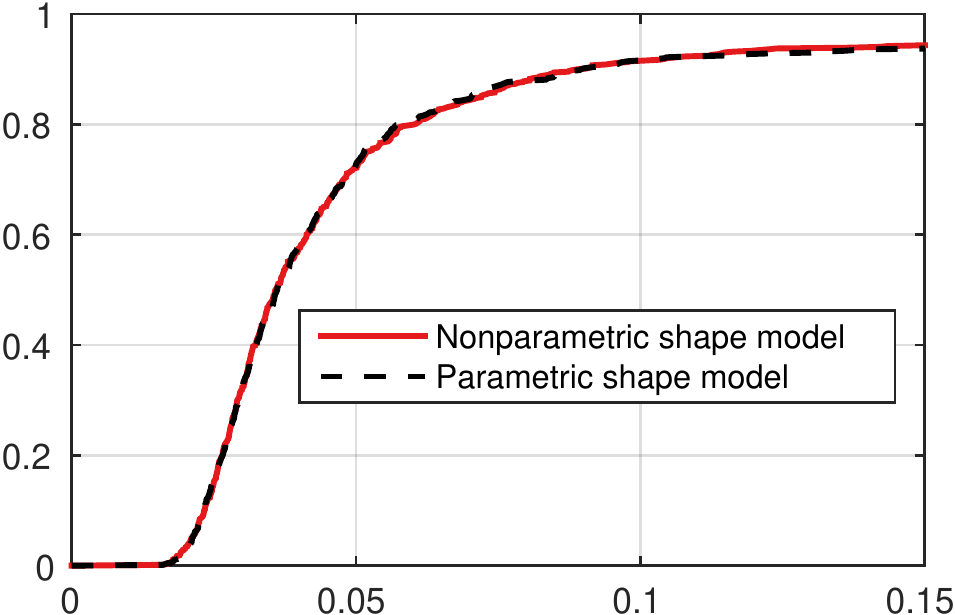}
\\
& Average normalized error \vspace{4pt}
\end{tabular}
\end{small}
\end{center}
\caption{
Comparison between two simplified version of our algorithm, identical except for the shape models used. 
The nonparametric version updates 2D landmark coordinates directly, while the parametric version updates PDM shape parameters. 
We see that accuracy is almost identical. 
} 
\label{fig-param_vs_non} %\vspace{-8pt}
\end{figure}

We address the first limitation by incorporating visibility into the regression. 
Specifically, we use global \textit{structured} point distribution models (SPDM) \cite{Rogers2001} to represent both shape variation and landmark visibility in a unified way, and we train regressors to predict SPDM parameters. 
SPDMs extend traditional point distribution models (PDMs) \cite{Cootes1995} by appending a ``structure'' vector (indicating whether each landmark is visible or not) to the set of shape parameters, and then constructing a combined linear model of shape and landmark visibility. 
This strategy separates us from the recent trend in face alignment of modeling shape nonparametrically (\eg, \cite{BurgosArtizzu2013,Cao2012,Kazemi2014,Ren2014,Wu2015,Xiong2013,Xiong2015,Yu2014}). 
We argue that PDMs have several desirable qualities:
\begin{itemize}
\item There are far fewer parameters to optimize, which results in a smaller (and faster to apply) set of regression coefficients. 
\item All landmarks are optimized simultaneously.
\item When learned via PCA, they generalize well to unfamiliar faces.
\end{itemize}
In fact, we show empirically in \fig{fig-param_vs_non} that there are no significant differences in accuracy and robustness between parametric and nonparametric shape models when used within otherwise identical systems.

We address the second limitation by introducing a \textit{branching} cascaded regression framework for face alignment. 
Conventional CSRs are \textit{single}-track; that is, they progress from one cascade level to the next in a straight line, with each regressor attempting to fit the entire dataset.
We instead split the regression problem into two or more simpler ones after each cascade level. 
Intuitively, each regressor can then operate on a simpler objective function (\ie, with fewer conflicting gradient directions). 
Our approach is inspired by Xiong and De la Torre \cite{Xiong2015} who introduced the concept of ``domains of homogeneous descent'' to define their Global Supervised Descent Method (GSDM). 
Their algorithm selects a \textit{single}-path CSR from a library of CSRs, each trained on a different region of the objective function domain, based on the result from the \textit{previous} frame in a video. 
In contrast, at runtime our algorithm chooses which branch to take, and thus which sequence of regressors to apply, \textit{based on the current test image}. 
Note that the method in \cite{Xiong2015} reverts to a conventional CSR on single images.

Most CSRs use off-the-shelf feature mapping functions (most commonly SIFT \cite{Lowe_IJCV2004_SIFT}) to produce feature descriptors from the image. 
Instead, we are inspired by the work of Ren \etal \cite{Ren2014} who showed that \textit{learned} feature mapping functions can be both extremely computationally efficient at runtime, and produce features that are task-specific. 
In \cite{Ren2014}, they used regression random forests \cite{Breiman2001,Criminisi2013} to induce local binary features that predict landmark updates. 
We propose to instead use regression random forests to induce local binary features that predict ideal SPDM parameter updates (\ie, for predicting both face shape and landmark visibility simultaneously) and to produce feature descriptors that are adaptive to landmark visibility. 

To summarize, we make the following contributions:
\begin{enumerate}
\item \textit{Branching cascaded regression:} We propose to divide the objective function domain after each cascaded regression level, which results in simpler regression problem at each subsequent cascade level. 
\item \textit{Profile-to-profile faces:} The proposed algorithm retains the state-of-the-art speed and accuracy of recent algorithms while handling a much wider range of head poses, including full profile faces with significant self-occlusion.  
\item \textit{Self-occlusion prediction:} To the best of our knowledge, we are the first to employ SPDMs to predict \textit{both} shape and landmark visibility within a cascaded shape regression framework. 
\item \textit{Nonparametric vs. parametric:} We show empirically (\fig{fig-param_vs_non}) that there is virtually no difference in accuracy between parametric and nonparametric shape models when used within otherwise identical systems on challenging faces.
\item \textit{Competitive results on challenging datasets:} The proposed approach is competitive with recent state-of-the-art methods on challenging frontal and near-frontal in-the-wild faces, and it produces excellent results on faces with significant non-frontal head pose.  
\item \textit{New in-the-wild dataset:} We have assembled and annotated a challenging dataset of in-the-wild profile faces from the Web to validate different parts of our algorithm and to promote future research in this area. 
\end{enumerate}

\section{Related Work}

Face alignment has a rich history in computer vision. % in computer vision. 
We forgo an extensive overview here due to space limitations and focus on the most relevant work. 

%Active Shape Models (ASMs) \cite{Cootes1995} and Active Appearance Models (AAMs) \cite{Cootes2001} were seminal works that inspired much subsequent investigation and improvement (\eg, \cite{Cootes2000,Cristinacce2007,Edwards1999,Matthews2004,Xiao2004} and many others). 
%ASMs model facial appearance locally and model facial shape globally using a Point Distribution Model (PDM). 
%PDMs represents shape as a linear combination of shape bases learned via Principal Component Analysis (PCA).
%AAMs use PDMs to model facial shape, but they model facial appearance holistically. 
%Constrained Local Models (CLMs) \cite{Saragih2009}, which build upon ASMs, came to conquer AAMs due to their improved generalizability and robustness in the wild. 
%CLMs use ``local experts'' (\eg, ) to produce local landmark responses, and use global shape models as regularizers to disambiguate local landmark estimates. 

Shape regression approaches have recently come to dominate the face alignment landscape \cite{BurgosArtizzu2013,Cao2012,Cao2013,Dollar2010,Jourabloo2015,Kazemi2014,Lee2015,Ren2014,Tzimiropoulos2015,Wu2015,Xiong2013,Yang2015}. 
Due to the complex relationships between image appearance and face shape updates, finding the true face shape in one step is difficult. 
Different strategies have been proposed for this problem, including trying multiple initializations and choosing the best result \cite{BurgosArtizzu2013,Cao2012}, and performing a coarse-to-fine search \cite{Zhu2015}. 
Another strategy, which we employ, is to split the regression problem into multiple iterations (\ie, a cascaded shape regressor (CSR)) and re-compute ``shape-indexed'' features \cite{Cao2012,Dollar2010} at each stage. 
This latter approach is perhaps best exemplified by the Supervised Descent Method (SDM) \cite{Xiong2013}. %, which inspired many subsequent methods. 

Conventional CSRs are \textit{single}-track; that is, they progress from one cascade level to the next in a straight line, with each regressor attempting to fit the entire dataset.
The problem with this strategy is that the objective function includes many conflicting gradient directions, especially when faces with a wide range of poses are included in the dataset. 
Xiong and De la Torre \cite{Xiong2015} proposed the Global Supervised Descent Method (GSDM) to address this problem.  
GSDM splits the objective function into regions of similar gradient directions and constructs a separate single-track CSR for each region. 
This results in a set of parallel, independent CSRs.
When used for face alignment, GSDM is restricted to video because it uses the result from the previous frame to select which CSR to use for the current frame. 
GSDM reverts to conventional CSR (\eg, SDM \cite{Xiong2013}) on single images. 
Instead, we introduce branching into the cascaded shape regression, which works on single images. 
During training, we split the regression problem into two or more simpler ones after each cascade level, as described in \sect{sec-approach}. 
This creates a tree-like cascade of shape regressors. 
At runtime our algorithm adaptively chooses which branch to take, and thus which sequence of regressors to apply, \textit{based on the current test image}. 
%Our algorithm works well on both video and single images. 

CSR-based methods can be divided into two categories: those that use off-the-shelf feature mapping functions like SIFT \cite{Lowe_IJCV2004_SIFT} 
(\eg, \cite{Tzimiropoulos2015,Xiong2013,Xiong2015}), and those that \textit{learn} feature mapping functions (\eg, using ensembles of regression trees \cite{Kazemi2014,Lee2015,Ren2014}). 
Learned feature mapping functions are task-specific, and are extremely computationally efficient at runtime when operating on simple pixel-differences \cite{Kazemi2014,Ren2014}. 
We use ensembles of regression trees in our approach to induce pixel-difference features correlated with our regression targets. 
Unlike \cite{Ren2014}, we novelly construct random regression forests with SPDM coefficients as targets (\sect{sec-approach}) instead of 2D offsets.

%Most CSR-based methods perform two steps at each regression level: feature extraction followed by shape regression. 
%Works that focus on the shape regression step (\eg, \cite{Tzimiropoulos2015,Xiong2013,Xiong2015}) tend to use off-the-self feature mapping functions like SIFT \cite{Lowe_IJCV2004_SIFT}. 
%An alternative is to \textit{learn} feature mapping functions \cite{Cao2012,Kazemi2014,Lee2015,Ren2014}. 
%Learned feature mapping functions are task-specific, and have been shown to be extremely computationally efficient at runtime \cite{Kazemi2014,Ren2014}.   
%Ferns \cite{Cao2012} and ensembles of regression trees \cite{Kazemi2014,Lee2015,Ren2014} have been proposed for this task. 
%Ferns or ensembles of regression trees are built up using a large number of ``shape-indexed'' (or ``pose-indexed'' \cite{Fleuret2008}) pixel-differences \cite{Cao2012,Kazemi2014,Ren2014} or difference of Gaussians \cite{Lee2015}. 
%Many features are sampled, and features that are best correlated with the regression target are selected. 
%For speed and robustness, we also use ensembles of regression trees and pixel-difference features, as described in \sect{}. 

Recent work \cite{Xiong2013,Xiong2015,Wu2015}, including those that learn feature mapping functions \cite{Cao2012,Kazemi2014,Ren2014}, have primarily used the 2D landmark coordinates directly as the regression targets. 
Instead we employ a PCA-based shape model and use the ideal shape parameter updates as regression targets. 
Parametric PCA-based models have several desirable qualities: they are very well studied, they require far fewer parameters than nonparametric shape models, they allow all landmarks to be optimized simultaneously, and they generalize well to unfamiliar faces. 
Tzimiropoulos showed that PCA-based 2D shape models are very competitive with nonparametric ones \cite{Tzimiropoulos2015}.
We show in \fig{fig-param_vs_non} that, empirically, there are no significant differences in accuracy and robustness between parametric and nonparametric shape models when used within otherwise identical systems.

%Shape-indexed features. Pixel-difference features specifically. 
%Cao \etal \cite{Cao2012}.
%Ren \etal \cite{Ren2014}.
%Fleuret and Geman \cite{Fleuret2008}.
%Kazemi and Sullivan \cite{Kazemi2014}.
%Lee \etal \cite{Lee2015}: shape-indexed difference of Gaussian (DoG) features. These are better features, but slightly slower to compute. 
%Overfitting becomes a problem, especially for shape-indexed features: they're very long, and they're coupled with the shape estimate: the shape estimate is determined by the shape-indexed features, and the shape-indexed features are extracted from the pixel coordinates references by the shape estimate -- from \cite{Lee2015}. 
%One approach: augment data: \cite{Cao2012,Kazemi2014,Lee2015,Ren2014}. 
%Another approach is ridge regression \cite{Ren2014,Tzimiropoulos2015}. 

Work that addresses significant head pose variation includes view-based models \cite{Cootes2000,Zhu2012}, nonlinear statistical models (\eg, kernel methods \cite{Romdhani1999} or mixture models \cite{Cootes1997,Kanaujia2006,Zhou2005}), and 3D shape models \cite{Baltrusaitis2012,Cao2013,Cao2014,Gu2006,Jourabloo2015,Matthews2007,Yu2013}. 
View-based models are straightforward, but require a separate model for each viewpoint, which is inefficient, and problems can arise at midpoints between models.
A key trait of our method is that it generates smoothly varying results across a wide range of viewpoints and facial expressions. 
View-based methods, including tree-structured face detectors \cite{Huang2007}, generally partition the training set in a discontinuous ad hoc way. 
In contrast, we partition training faces into overlapping sets in a principled way using PCA. 
Our method is much faster than view-based methods, which typically compute and score multiple candidate results for each face.

Nonlinear statistical models tend to be too slow for real-time applications. 
One practical problem with 3D models is that publicly available training data are limited, and most 3D algorithms focus on either easier in-the-lab face datasets \cite{Baltrusaitis2012,Cao2013,Cao2014,Matthews2007} and/or near fronto-parallel faces \cite{Cao2013,Cao2014,Yu2013}. 
Another interesting approach is \textit{dense} 3D face alignment and reconstruction (\eg, \cite{Jeni2015} and its predecessors), although results suggest that test faces must still be near-frontal.

%Xiong and De la Torre \cite{Xiong2015} addressed the problem of producing regressors for nonlinear least squares problems in computer vision, including multi-view face tracking. 
%They defined ``domains of homogeneous descent'' and proved their existence as part of their Global Supervised Descent Method (GSDM). 
%GSDM (as proposed) requires a video when applied to face alignment. 
%The strategy is to use the shape update result from the previous frame to select a \textit{single}-path CSR from a library of CSRs, each trained on a different region of the objective function domain, for the current frame. 
%On a single image, GSDM reverts to a conventional CSR. 
%In contrast, we split the regression problem into two or more simpler ones after each cascade level. 
%At runtime our algorithm adaptively chooses which branch to take, and thus which sequence of regressors to apply, \textit{based on the current test image alone}. 
%GSDM does not address landmark occlusion. 

%Self-occlusion of landmarks becomes increasingly significant as faces rotate away from frontal. 
Until very recently, most research on landmark occlusion has focused on near-frontal faces \cite{BurgosArtizzu2013,Yu2014}. %, which is a very challenging problem. 
%Estimating the occlusion state of landmarks on near-frontal faces is extremely challenging because 
The only cue is appearance (face shape is generally uncorrelated), and the presence of occluders is arbitrary. %can be anything. %is relatively unpredictable (an occluder can be anything). 
We instead focus on the more tractable problem of estimating occlusion due solely to head pose, which is highly correlated with 2D face shape. 
We capitalize on this fact by modeling shape and visibility simultaneously using SPDMs \cite{Rogers2001}, as described in \sect{sec-approach}. 
Note that the original SPDM authors did not propose using SPDMs in this way.

Very recently, researchers have considered the problem of locating landmarks in faces with significant head rotation and occlusion within the CSR framework. 
Wu and Ji \cite{Wu2015} proposed an approach for estimating arbitrary landmark visibility. 
%At each cascade level, their algorithm updates the face shape and the visibility of each landmark. 
%Their landmark visibility prediction model includes a loss function that penalizes ``infrequent and infeasible occlusion label configurations.'' 
Inference of landmark visibility with their model is nontrivial, and they used Monte Carlo approximation, which is relatively slow. 
Instead, we estimate pose-related landmark visibility in a simpler and much faster way. 
Like our approach, their shape regressors take landmark visibility into account. % so that predictors rely more on features extracted at visible landmarks. 
%Likewise, our shape regressors are adaptive to landmark visibility. 
%We do not address arbitrary landmark visibility. 

Jourabloo and Liu \cite{Jourabloo2015} proposed learning a cascaded coupled regressor where, at each cascade level, they learn two regressors: one to estimate parametric 3D shape updates and one to estimate the parameters of a weak perspective model. 
They estimate landmark visibility indirectly by first estimating the surface normal of each landmark, and then labeling landmarks with negative $z$ coordinates as invisible. 
In this work, we incorporate pose-related landmark visibility into our model in a unified way, % using Structured Point Distribution Models (SPDMs) \cite{Rogers2001}, 
which allows our algorithm to estimate shape and landmark visibility simultaneously. 
Like previous CSRs, \cite{Jourabloo2015} and  \cite{Wu2015} are single-path, whereas our algorithm chooses the cascade path at runtime. 
%Interestingly, \cite{Jourabloo2015} requires at least $10$ cascade levels for linear regression, and $75$ for their fern regressors, whereas our algorithm requires only $4$ to achieve good performance.

\section{Our Approach} \label{sec-approach}

We beginning with a brief overview of Structured Point Distribution Models (SPDMs) \cite{Rogers2001}, followed by a description of our approach. 

\subsection{Modeling Landmark Visibility with SPDMs}

To build an SPDM we first build a Point Distribution Model (PDM) \cite{Cootes1995}. 
PDMs model a set of $n$ shapes $\shapesMat = [\shape_{1}, \cdots, \shape_{n}]$ using a linear combination of shape bases $\shapeBasesMat_{\shapesMatSymb}$ plus the mean shape  $\mathbf{\mu}_{\shapesMatSymb}$:
\begin{equation} \label{eq-pdm}
\hat{\shapesMat} = \mathbf{\mu}_{\shapesMatSymb} + \shapeBasesMat_{\shapesMatSymb} \shapeParamVec_{\shapesMatSymb}, \ \ \ \ \ \shapesMat \approx \hat{\shapesMat}, 
\end{equation}
where $\shapeParamVec_{\shapesMatSymb}$ contains the reconstruction parameters; 
$\mathbf{\mu}_{\shapesMatSymb}$, $\shapeBasesMat_{\shapesMatSymb}$, and $\shapeParamVec_{\shapesMatSymb}$ are computed from $\shapesMat$ via PCA.

Missing landmarks in $\shapesMat$ must be imputed in order to compute $\shapeBasesMat_{\shapesMatSymb}$ and $\shapeParamVec_{\shapesMatSymb}$. 
Several imputation strategies exist, but we employ Iterated PCA as described in \cite{Rogers2001}. 
The missing landmarks in $\shapesMat$ are initialized using mean value imputation, \ie, $\pnt_{\pntIndex} = \hat{\pnt}_{\pntIndex} = \mu_{\pntIndex}$, where $\pntIndex$ indexes missing landmarks. 
We then set the number of bases to $\numBases = 1$ and cycle between computing $\{\mathbf{\mu}_{\shapesMat}$, $\shapeBasesMat_{\shapesMatSymb,\numBases}$, $\shapeParamVec_{\shapesMatSymb,\numBases}\}$ and setting $\pnt_{\pntIndex} = \hat{\pnt}_{\pntIndex}$ for all $\pntIndex$ until convergence. 
$\numBases$ is then incremented by $1$ and the process is repeated starting with the previous result. 
We inspect the imputed results after each loop and continue incrementing $\numBases$ until we see no further improvement. 
We note that imputed landmarks need not be perfect since their visibility will be set to zero later.
Once imputation is complete, we can compute $\shapeParamVec_{\shapesMatSymb}$ in the usual way. 

Prior to PCA, the shapes in $\shapesMat$ are aligned using Procrustes Analysis to remove variations due to global face rotation, translation, and scale. 
To incorporate these variations back into the model, we append four additional basis vectors to $\shapeBasesMat_{\shapesMatSymb,\numBases}$, one for  rotation, one for x-translation, one for y-translation, and one for scale, as described in Section 4.2.1 of \cite{Matthews2004}, and re-orthonormalize.

We augment each shape with a binary ``structure'' vector $\visVec$ that indicates whether each landmark is visible ($1$) or not ($0$).
%Like $\shapesMat$, we form a matrix $\visMat$ composed of the structure vectors. 
PCA on $\visMat = [\visVec_{1}, \cdots, \visVec_{n}]$ gives a matrix $\shapeBasesMat_{\visVec}$ of visibility bases and a matrix $\shapeParamVec_{\visVec}$ of continuous landmark visibility parameters. 
The training shapes can then be represented by a combined set of parameters:
\begin{equation}
\shapeParamVec = \left[ \begin{array}{c} \shapeParamVec_{\shapesMatSymb} \\ \shapeParamVec_{\visVec}\end{array} \right].
\end{equation}
%$\shapeParamVec = \transp{(\transp{\shapeParamVec_{\shapesMat}}, \transp{\shapeParamVec_{\visMat}})}$. 
Finally, a combined model of shape and visibility is constructed by performing PCA once more on the combined parameters: $\shapeParamVec = \combBasesMat \combParamVec$, where $\combBasesMat$ expresses correlations between shape and visibility, and $\combParamVec$ is a new parameter space. 
Note that it is important to first shift and scale $\shapeParamVec_{\shapesMatSymb}$ so that its elements lie in the same range as $\shapeParamVec_{\visVec}$. 
See \cite{Rogers2001} for more details. 

\subsection{Overview}

We first give a brief overview of our approach and then present the technical details of each step. 
Our branching cascaded regressor (BCR) is a binary tree that takes a face image $\image$ and a rough initial shape estimate $\shape^{0}$ as input, and produces a final shape estimate $\shape^{\numCascadeLevels}$ as output. 
Initially, all landmarks are assumed to be visible (\ie, $\visVec = \mathbf{1}$). 
$\shape$ and $\visVec$ are progressively updated as the input face traverses the tree. 
%, and one of the leaf note at level $\numCascadeLevels$ produces the final shape estimate $\shape^{\numCascadeLevels}$. 
At runtime, each BCR node performs two tasks: (1) a shape and landmark visibility update followed by (2) a binary classification of the input face. 
Depending on the classifier's output, the face is sent to either the left child or the right child for the next cascade level.
Each child node represents a different mode of regression targets. 
The classifier aims to send the input face to the child node that will best predict its shape. 

%During training, the training examples are partitioned into left and right branches such that the child nodes 

%split into two overlapping subsets, one for each child node. 
%The subsets represent two different modes of regression targets. 
%Intuitively, this makes the regression problem easier in the next cascade level. 

%We explain this idea in more detail in \sect{}. 
%First, we describe 

%Specifically, we perform dimensionality reduction on the ideal shape and visibility updates via PCA, project the data onto the first (most dominant) subspace dimension, and partition the training set. 
%We train a linear classifier 

\subsection{Shape and Landmark Visibility Regression} \label{sec-shape_vis_reg}

The shape and visibility update at cascade level $\cascadeIndex$ in BCR node $\bcrNodeIndex$ is performed by $\shapeRegressor^{\cascadeIndex,\bcrNodeIndex}$. 
We omit $\bcrNodeIndex$ below for simplicity, but it is important to note that each BCR node has its own shape regressor and SPDM. 
Each BCR node models a subset of similar training faces, which allows each shape regressor and SPDM to be much simpler (fewer parameters) than a single global model. 
The input to regressor $\cascadeIndex$ is a feature descriptor $\featDesc(\image, \shape)$ that captures the local appearance in image $I$ relative to shape $\shape$. 
The output is an updated shape and visibility estimate: 
\begin{eqnarray}
\Delta \combParamVec^{\cascadeIndex} &=& \shapeRegressor^{\cascadeIndex} \featDesc^{\cascadeIndex}(\image, \shape^{\cascadeIndex-1}) \\
\combParamVec^{\cascadeIndex} &=& \combParamVec^{\cascadeIndex-1} + \Delta \combParamVec^{\cascadeIndex} \\
\left[ \begin{array}{c} \shapeParamVec_{\shapesMatSymb}^{\cascadeIndex} \\ \shapeParamVec_{\visVec}^{\cascadeIndex} \end{array} \right] &=& \combBasesMat^{\cascadeIndex} \combParamVec^{\cascadeIndex} \\
\shape^{\cascadeIndex} &=& \mathbf{\mu}_{\shapesMatSymb}^{\cascadeIndex} + \shapeBasesMat_{\shapesMatSymb}^{\cascadeIndex} \shapeParamVec_{\shapesMatSymb}^{\cascadeIndex} \\
\visVec^{\cascadeIndex} &=& \mathbf{\mu}_{\visVec}^{\cascadeIndex} + \shapeBasesMat_{\visVec}^{\cascadeIndex} \shapeParamVec_{\visVec}^{\cascadeIndex}
\end{eqnarray}
Note that $\combBasesMat$, $\shapeBasesMat_{\shapesMatSymb}$, and $\shapeBasesMat_{\visVec}$ are specific to each $\{\cascadeIndex,\bcrNodeIndex\}$, and so we recompute $\combParamVec^{\cascadeIndex-1}$ from $\shape^{\cascadeIndex-1}$ and $\visVec^{\cascadeIndex-1}$ at each level before computing $\combParamVec^{\cascadeIndex}$:
\begin{eqnarray}
\shapeParamVec_{\shapesMatSymb}^{\cascadeIndex-1} &=& \transp{\shapeBasesMat_{\shapesMatSymb}^{\cascadeIndex}} (\shape^{\cascadeIndex-1} - \mathbf{\mu}_{\shapesMatSymb}^{\cascadeIndex}) \\
\shapeParamVec_{\visVec}^{\cascadeIndex-1} &=& \transp{\shapeBasesMat_{\visVec}^{\cascadeIndex}} (\visVec^{\cascadeIndex-1} - \mathbf{\mu}_{\visVec}^{\cascadeIndex}) \\
\combParamVec^{\cascadeIndex-1} &=& \transp{\combBasesMat^{\cascadeIndex}} \left[ \begin{array}{c} \shapeParamVec_{\shapesMatSymb}^{\cascadeIndex-1} \\ \shapeParamVec_{\visVec}^{\cascadeIndex-1} \end{array} \right] 
\end{eqnarray}
Each $\shapeRegressor^{\cascadeIndex}$ is computed offline by solving the following ridge regression problem:
\begin{equation}\label{eq-Robj}
\argmin_{\shapeRegressor^{\cascadeIndex}} \sum_{\imgIndex \in \nodeMems_{\cascadeIndex}} \| \Delta \hat{\combParamVec}_{\imgIndex}^{\cascadeIndex} - \shapeRegressor^{\cascadeIndex} \featDesc^{\cascadeIndex}(\image_{\imgIndex}, \shape^{\cascadeIndex-1})\|_{2}^{2} + \lambda \| \shapeRegressor^{\cascadeIndex} \|_{2}^{2},
\end{equation}
where $\Delta \hat{\combParamVec}_{\imgIndex}^{\cascadeIndex}$ is the ideal parameter update for face $\imgIndex$ and $\nodeMems_{\cascadeIndex}$ are the training faces that belong to the current BCR node. 
Ridge regression (\ie, the second regularization term) is necessary because $\featDesc^{\cascadeIndex}(\image, \shape)$ is very high dimensional (see \sect{sec-learning_feat_map_func}) and substantial overfitting would result without it \cite{Ren2014}. 

\subsection{Branching}

\begin{figure}
\begin{center}
\begin{small}
\begin{tabular}{ccl}
\includegraphics[height=1.0in]{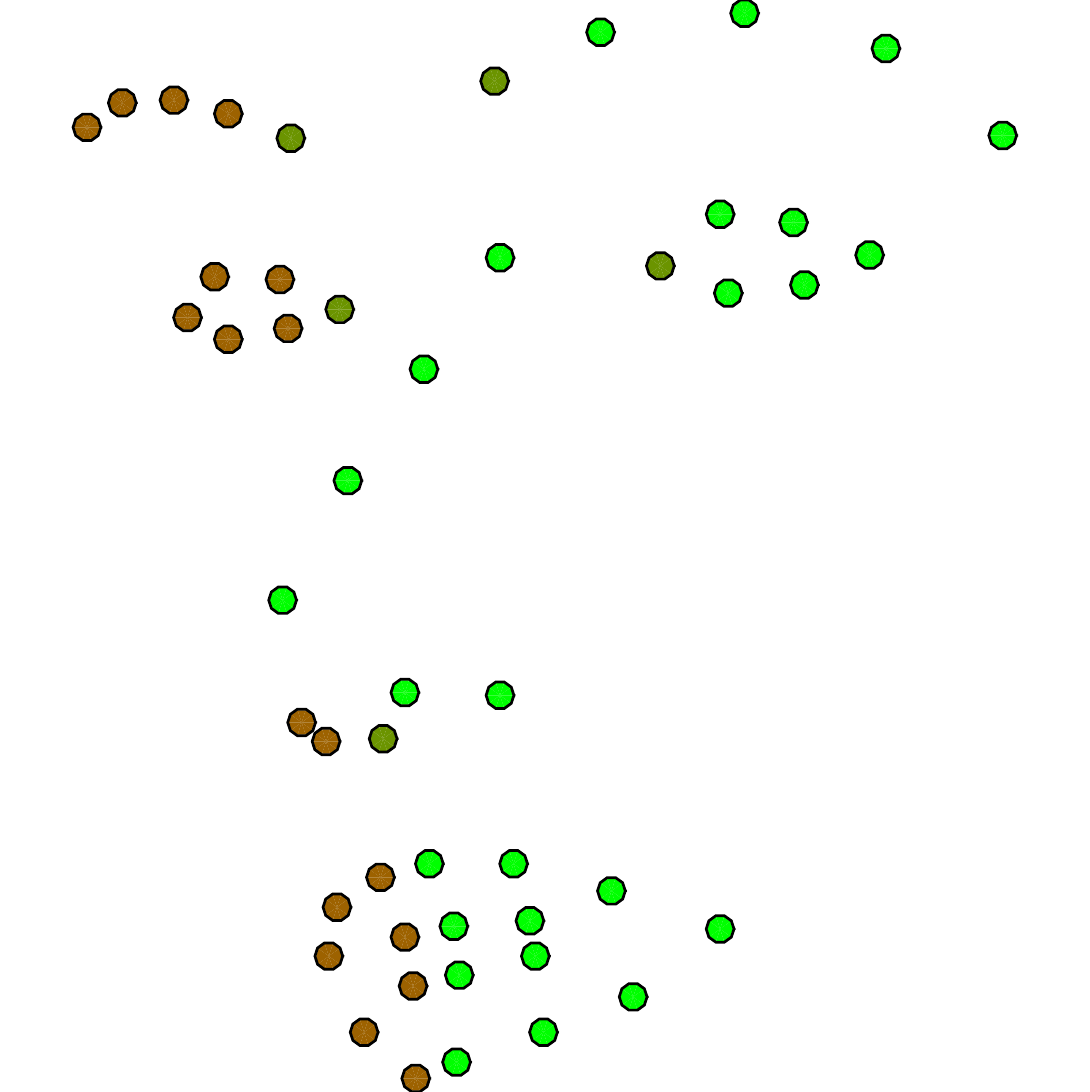}
&
\includegraphics[height=1.0in]{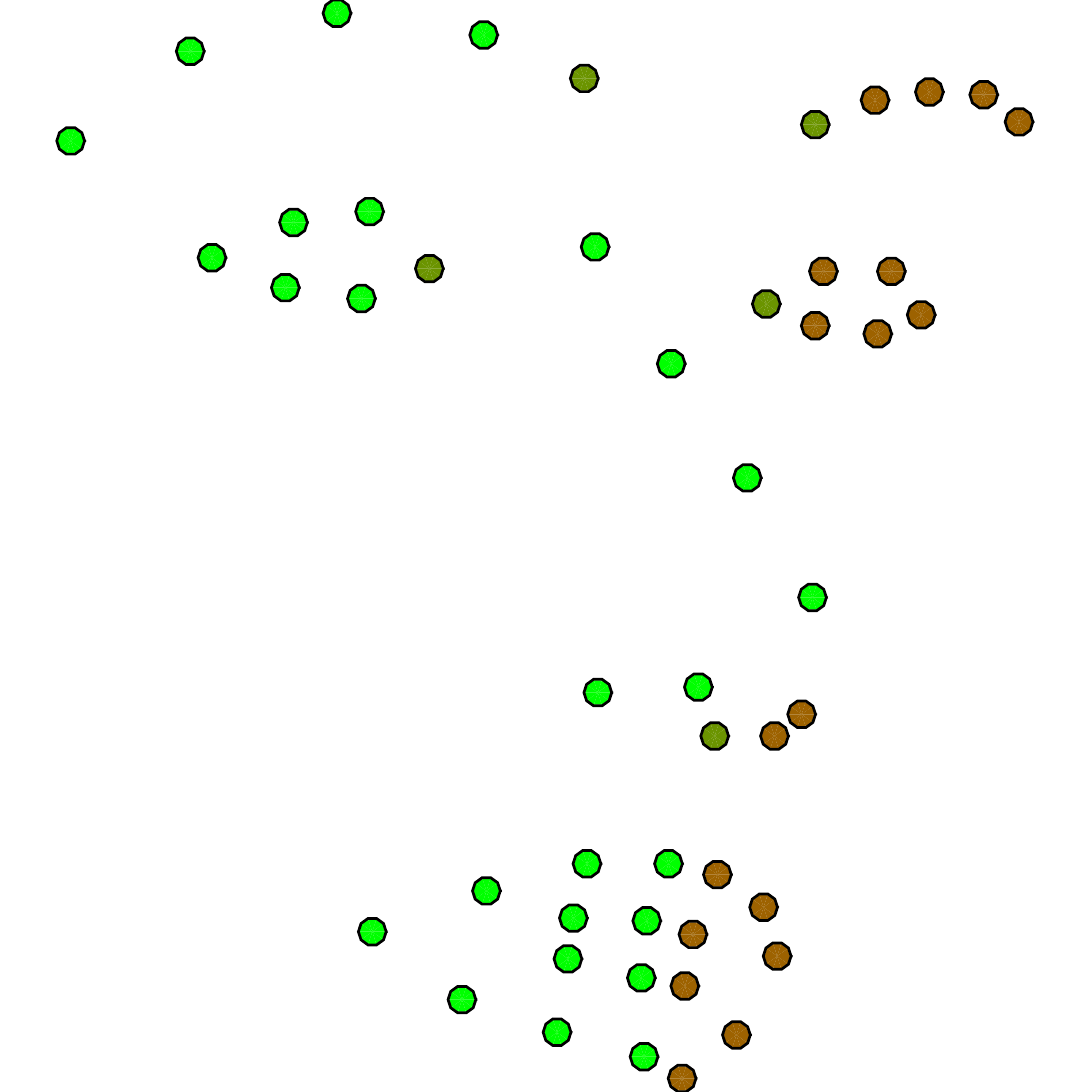}
&
\includegraphics[height=1.0in]{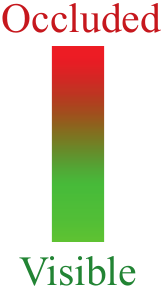} 
\end{tabular}
\end{small}
\end{center}
\caption{
Two faces generated by varying the first dimension of our combined shape and landmark visibility model in the first level of the cascade. 
Note that yaw head rotation is naturally encoded, and the visibility of landmarks shifts smoothly as they rotate toward being occluded.  
} 
\label{fig-modes_of_var} %\vspace{-8pt}
\end{figure}

During training, the training examples traverse the BCR much like test faces do. 
After learning $\shapeRegressor^{\cascadeIndex,\bcrNodeIndex}$ in each node, the training examples are partitioned into two overlapping sets, one for each child. 
The goal is to cluster the training examples into the two most significant modes of regression targets. 
Intuitively, this gives each child node a regression problem with a simpler objective function.  
Specifically, we perform dimensionality reduction on the ideal shape and landmark visibility updates via PCA, project the data onto the first (most dominant) subspace dimension, and partition the training set along that dimension.
This is equivalent to partitioning along the first dimension of $\Delta \hat{\combParamVec}_{\imgIndex}^{\cascadeIndex}$ in \eq{eq-Robj}. 
\fig{fig-modes_of_var} shows two shapes generated by varying this first mode, which naturally encodes both yaw head rotation and pose-related landmark visibility.  
 
We train an L2-regularized linear SVM \cite{Fan2008} to predict partition labels from feature descriptors $\featDesc(\image, \shape^{\cascadeIndex-1})$.
At test time, our algorithm determines which direction to branch based on the SVM output:
\begin{equation}
y^{\cascadeIndex} = \transp{\mathbf{w^{\cascadeIndex}}} \featDesc^{\cascadeIndex}(\image, \shape^{\cascadeIndex-1}) + b^{\cascadeIndex}, 
\end{equation}
(\eg, $y < 0$ results in a left child traversal). 
%Note that this operation is extremely fast. %, especially when no more expensive than a single landmark coordinate update. 

Due to image noise, poor initialization, \etc, mistakes are inevitable, especially near the decision boundary. 
Therefore, \textit{we retain a sizable fraction of the most difficult-to-classify training faces in each child node that belong to the other branch.}
This way, if test faces branch incorrectly, subsequent regressors will be general enough to handle them. 
In our implementation we simply send all training examples down both branches, and throw away the easiest-to-classify $1/3$ that belong to the other branch. 
In the first node, for example, this results in a $99.5\%$ recall rate for training instances. 

% Feature mapping function
\subsection{Learning Feature Mapping Functions} \label{sec-learning_feat_map_func}

In training, each $\featDesc^{\cascadeIndex}(\image, \shape)$ is learned separately for each BCR node prior to learning each $\shapeRegressor^{\cascadeIndex}$. 
$\featDesc^{\cascadeIndex}$ is composed of many \textit{local} independent feature mapping functions: $\featDesc^{\cascadeIndex} = [\littleFeat_{1}^{\cascadeIndex}, \cdots, \littleFeat_{\numTrees}^{\cascadeIndex}]$. 
%Each $\littleFeat_{\treeIndex}^{\cascadeIndex}$ aims to minimize the following objective function:
%\begin{equation}
%\argmin_{\littleRegressor^{\cascadeIndex}, \littleFeat_{\treeIndex}^{\cascadeIndex}} \sum_{\imgIndex \in \nodeMems_{\cascadeIndex}} \| \Delta \hat{\combParamVec}_{\imgIndex}^{\cascadeIndex} - \littleRegressor^{\cascadeIndex} \littleFeat_{\treeIndex}^{\cascadeIndex}(\image_{\imgIndex}, \shape^{\cascadeIndex-1})\|_{2}^{2}. 
%\end{equation}
%Each $\littleFeat_{1}^{\cascadeIndex}$ is centered on a landmark estimate. 

We use an ensemble of regression trees \cite{Breiman2001} to learn each $\littleFeat_{\treeIndex}^{\cascadeIndex}$ according to \cite{Ren2014}, with several key differences. 
First, instead of using the 2D landmark coordinates as regression targets, we use SPDM parameters. 
Second, instead of allocating trees uniformly across landmarks, we allocate them proportionally to the mean visibility of each landmark in the training data associated with each BCR node. 
For BCR nodes in which profile faces dominate, for example, the result is an ensemble of trees that emphasize visible landmarks. 
Third, we train the split nodes of each tree in a different way. 

To train each split node, our algorithm first chooses one dimension of $\Delta \hat{\combParamVec}_{\imgIndex}^{\cascadeIndex}$. 
The choice is random, but weighted, where the weight is proportional to the amount of variance that each dimension of $\Delta \hat{\combParamVec}_{\imgIndex}^{\cascadeIndex}$ encodes. 
This emphasizes the most significant modes of variation in the SPDM.  
We then test $\numRandFeats = 500$ random pixel-difference features \cite{Ren2014} and select the one that gives rise to maximum variance reduction in the chosen dimension of $\Delta \hat{\combParamVec}_{\imgIndex}^{\cascadeIndex}$.

\begin{figure*}
\begin{center}
\begin{small}
\begin{tabular}{c@{ }c@{ }c@{ }c@{ }c@{ }c@{ }c@{ }c}
\includegraphics[width=0.82in]{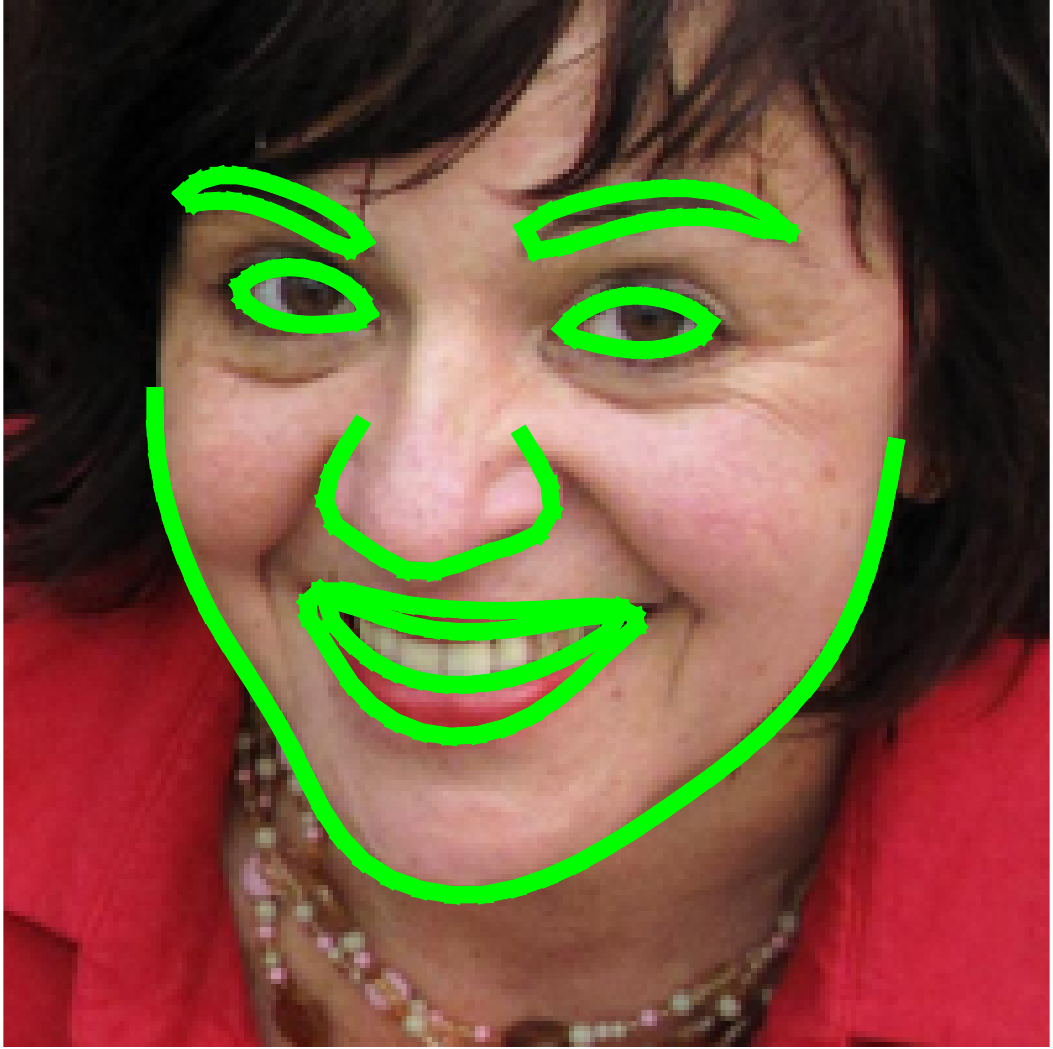}
&
\includegraphics[width=0.82in]{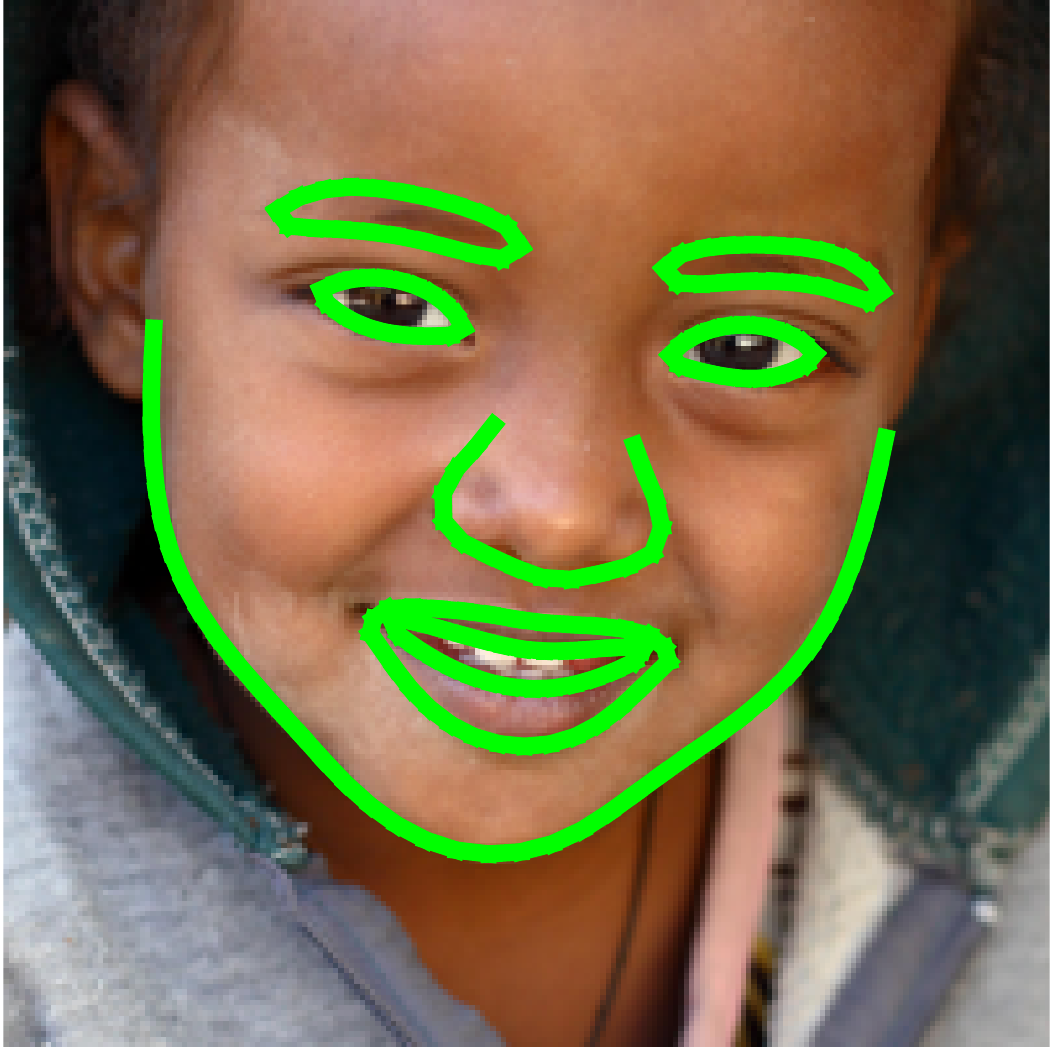}
&
\includegraphics[width=0.82in]{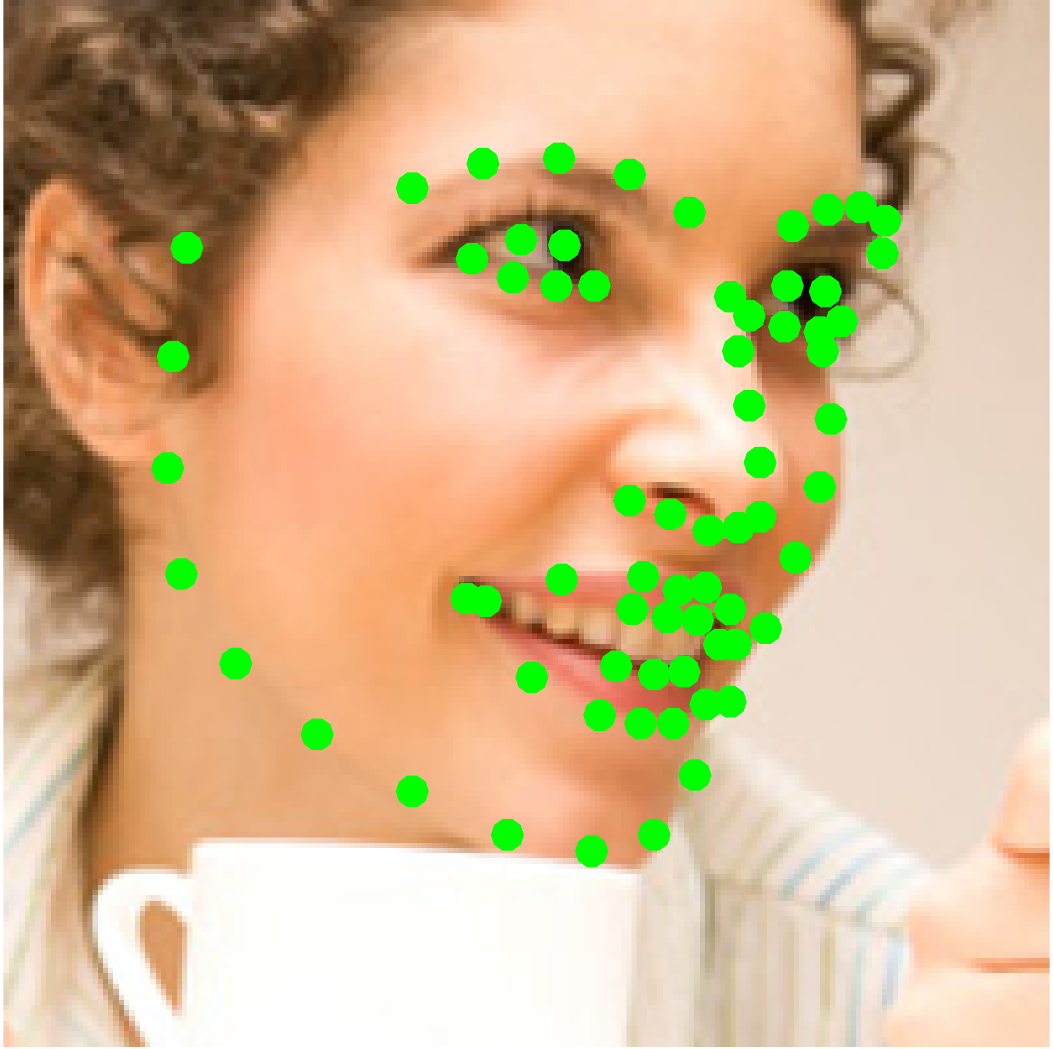}
&
\includegraphics[width=0.82in]{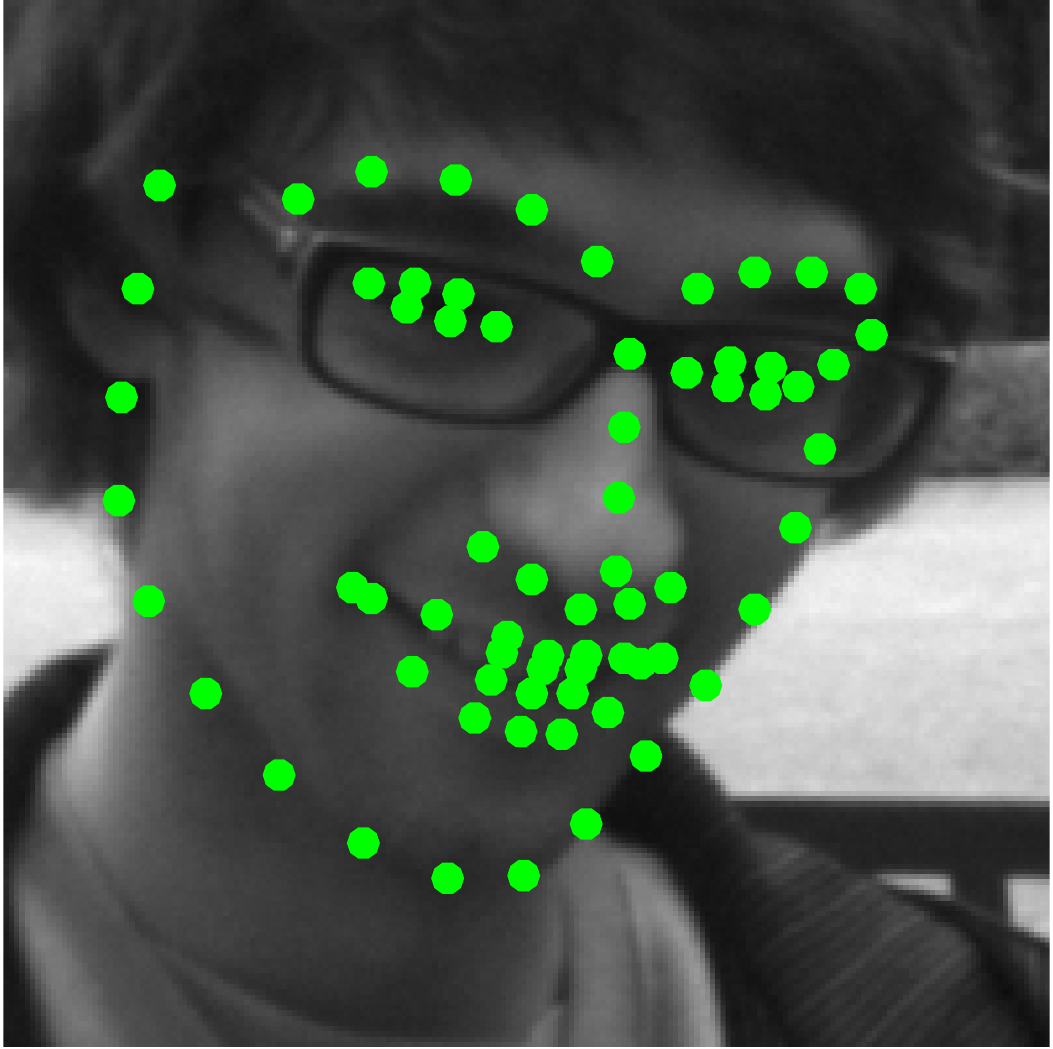}
&
\includegraphics[width=0.82in]{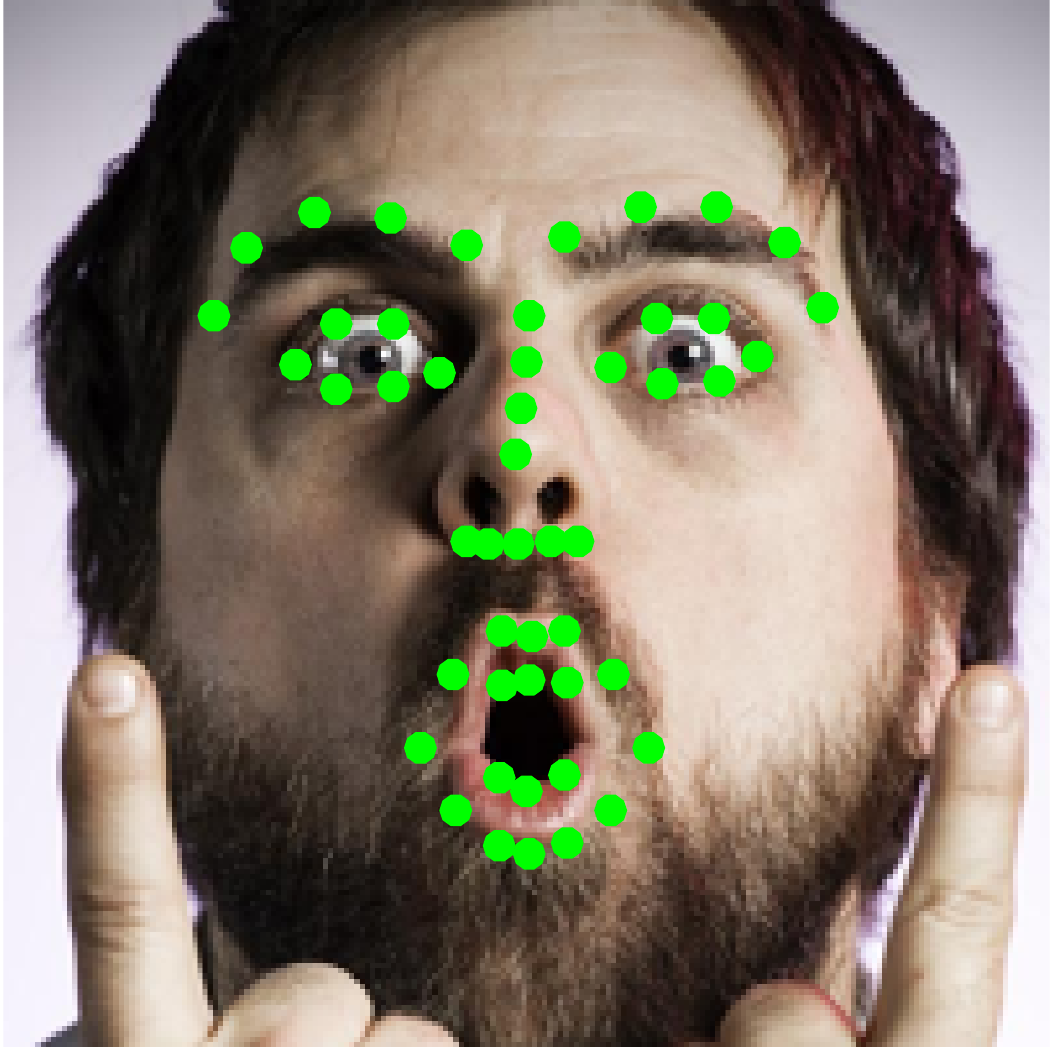}
&
\includegraphics[width=0.82in]{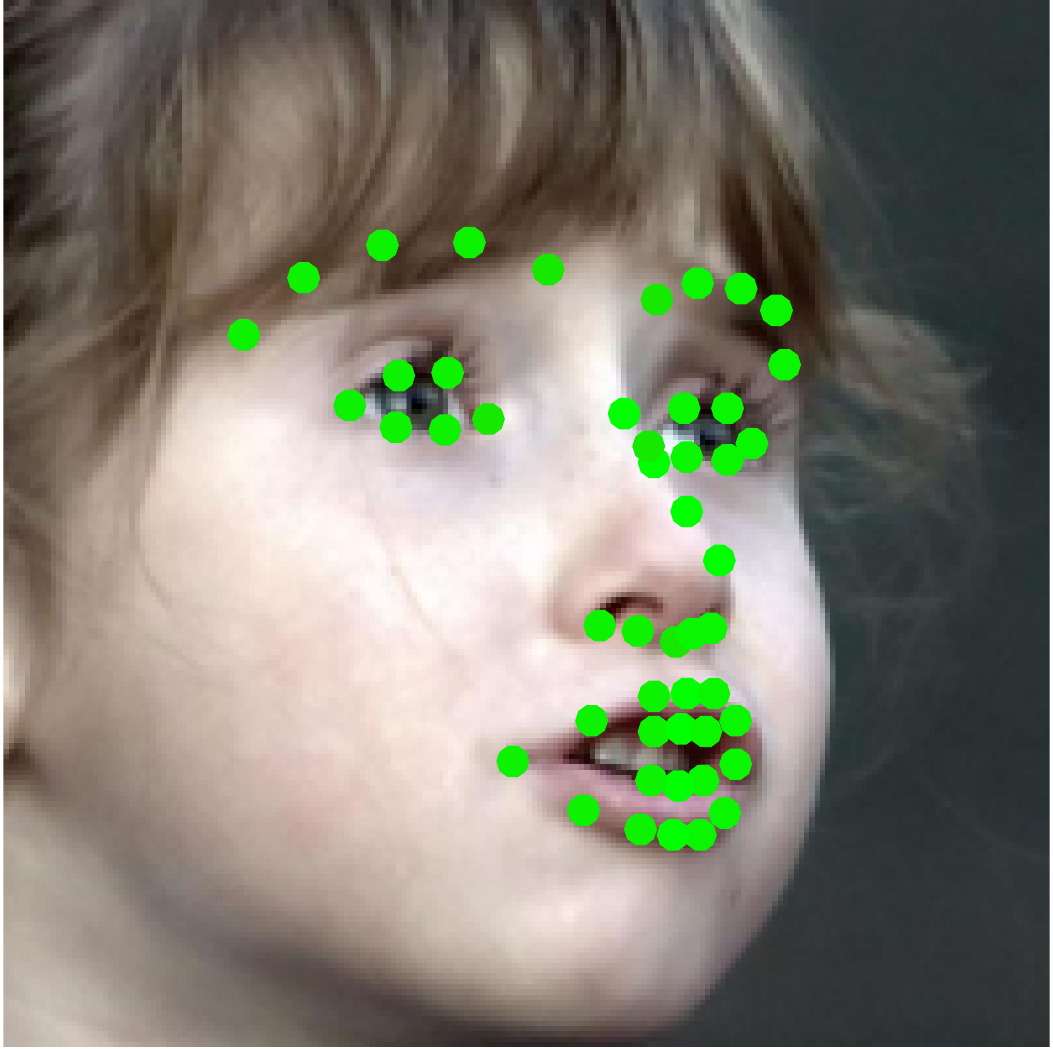}
&
\includegraphics[width=0.82in]{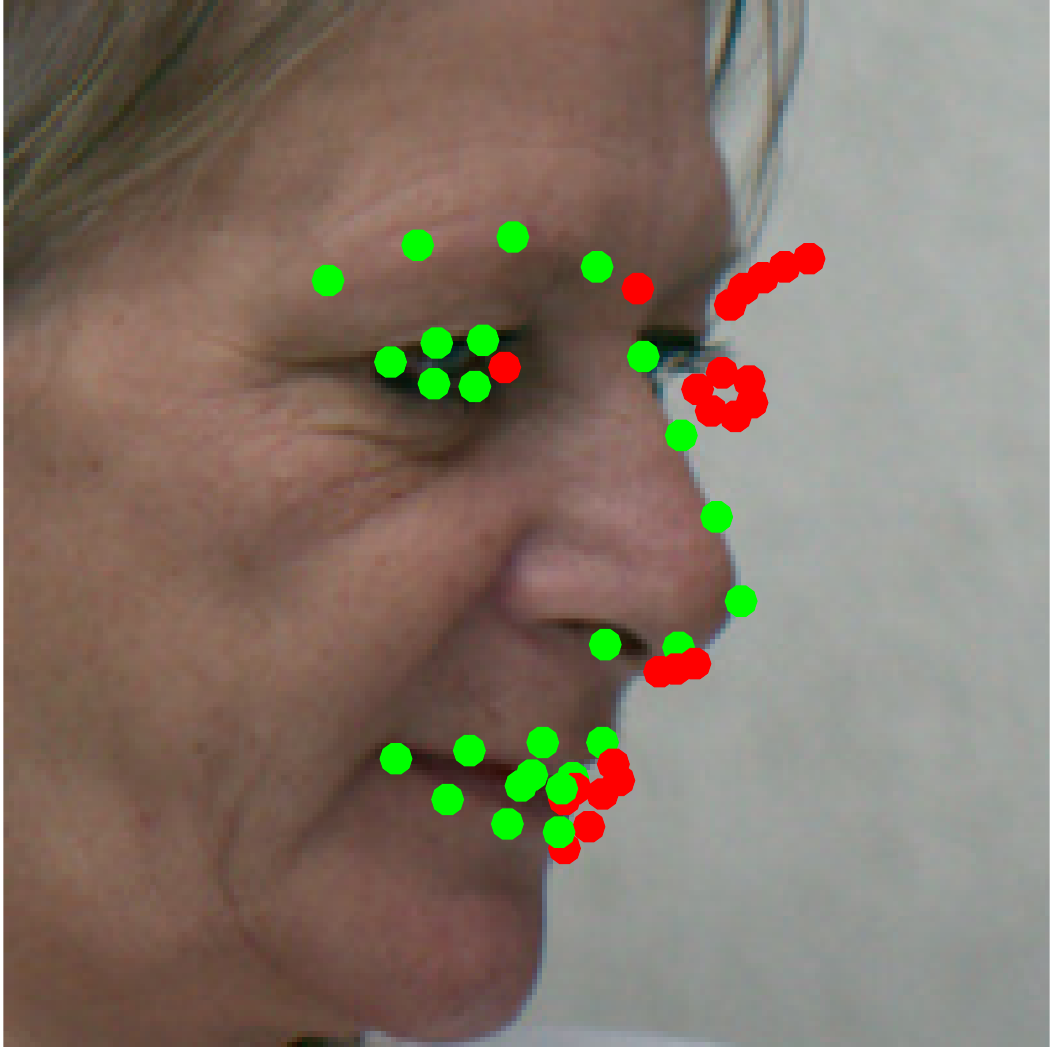}
&
\includegraphics[height=0.82in]{vis_occ_key.pdf}
\\
\includegraphics[width=0.82in]{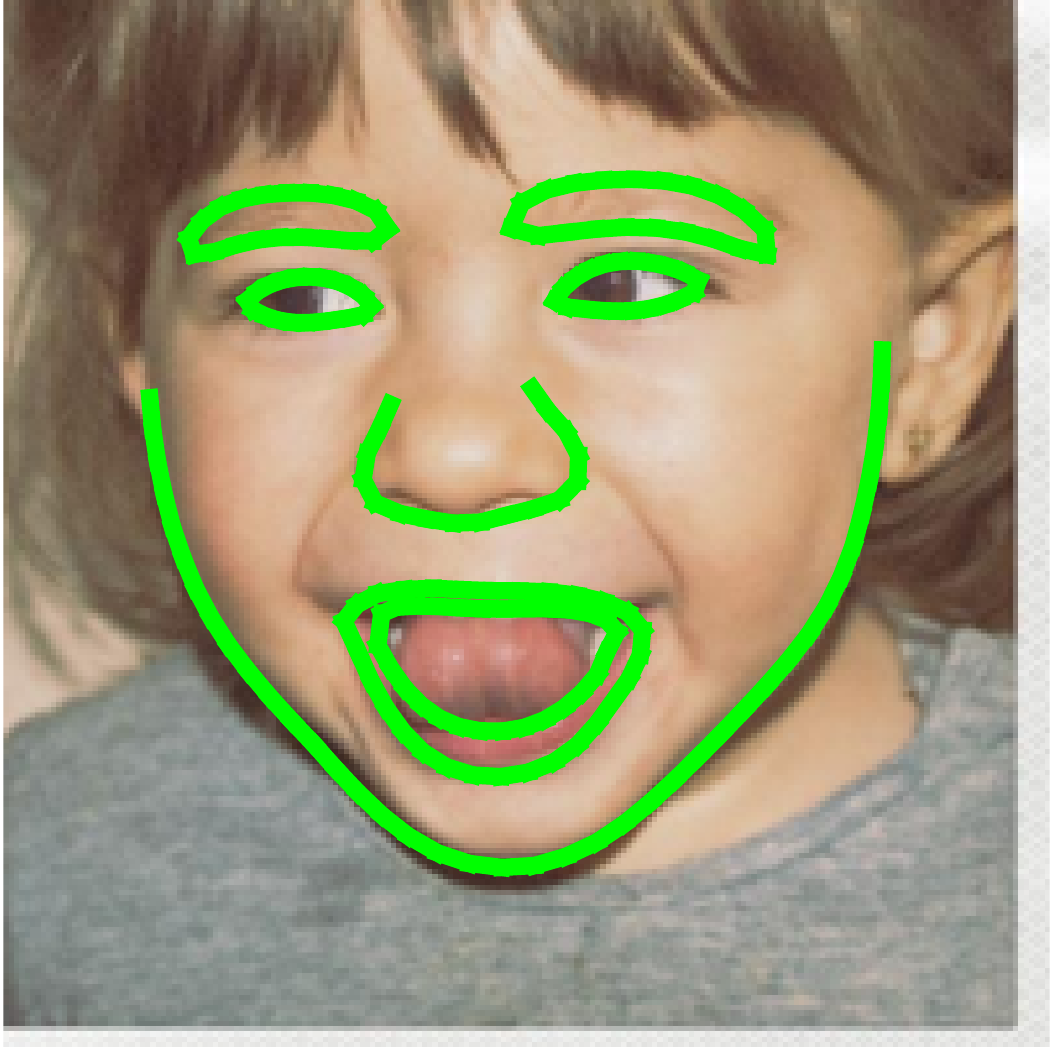}
&
\includegraphics[width=0.82in]{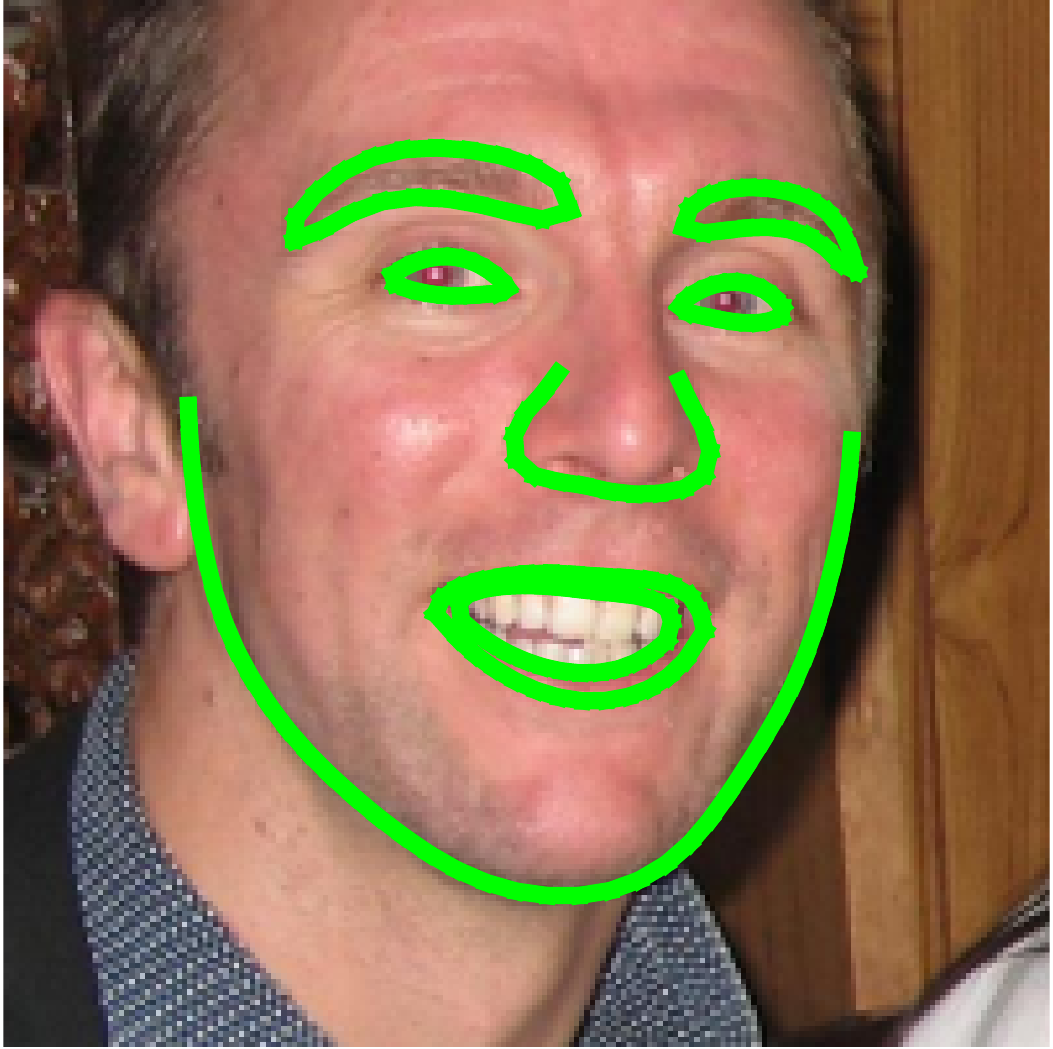}
&
\includegraphics[width=0.82in]{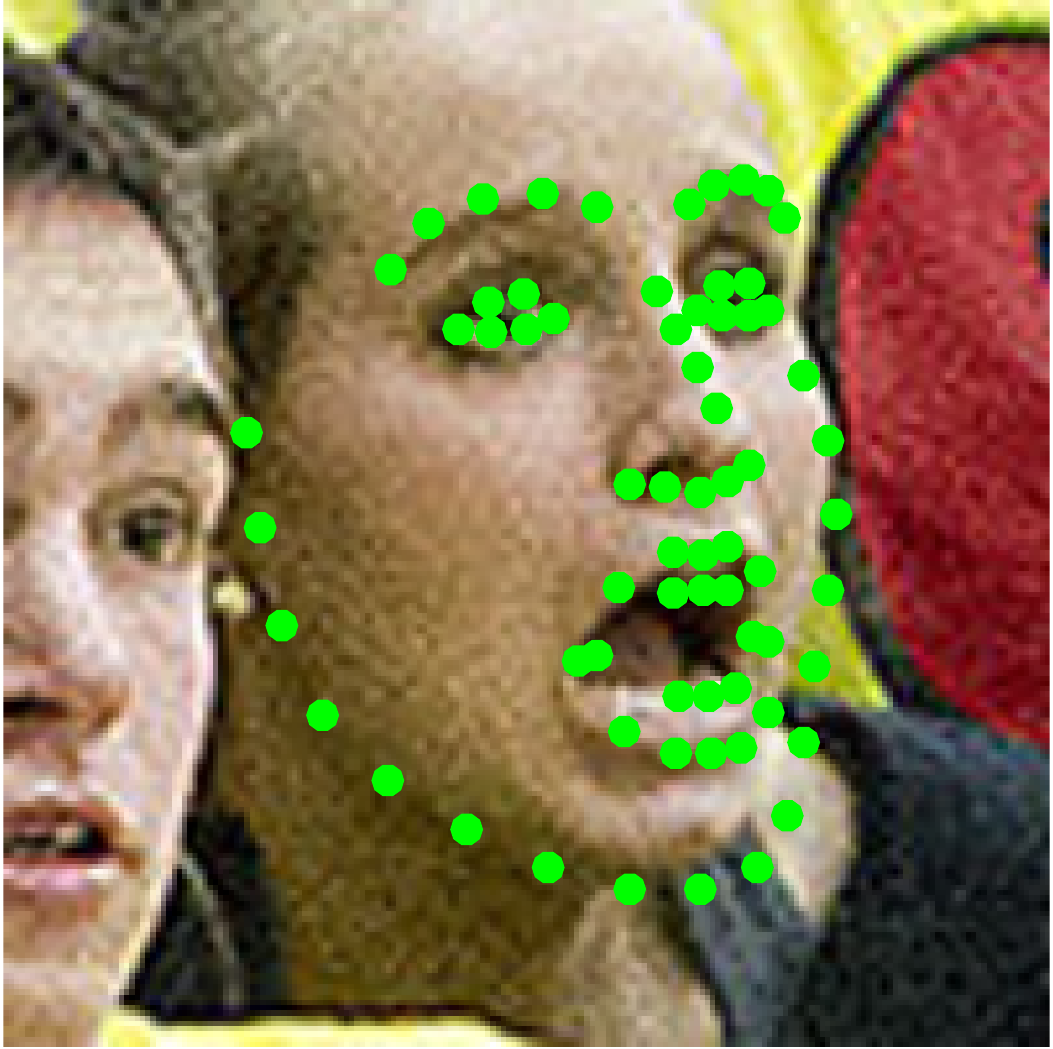}
&
\includegraphics[width=0.82in]{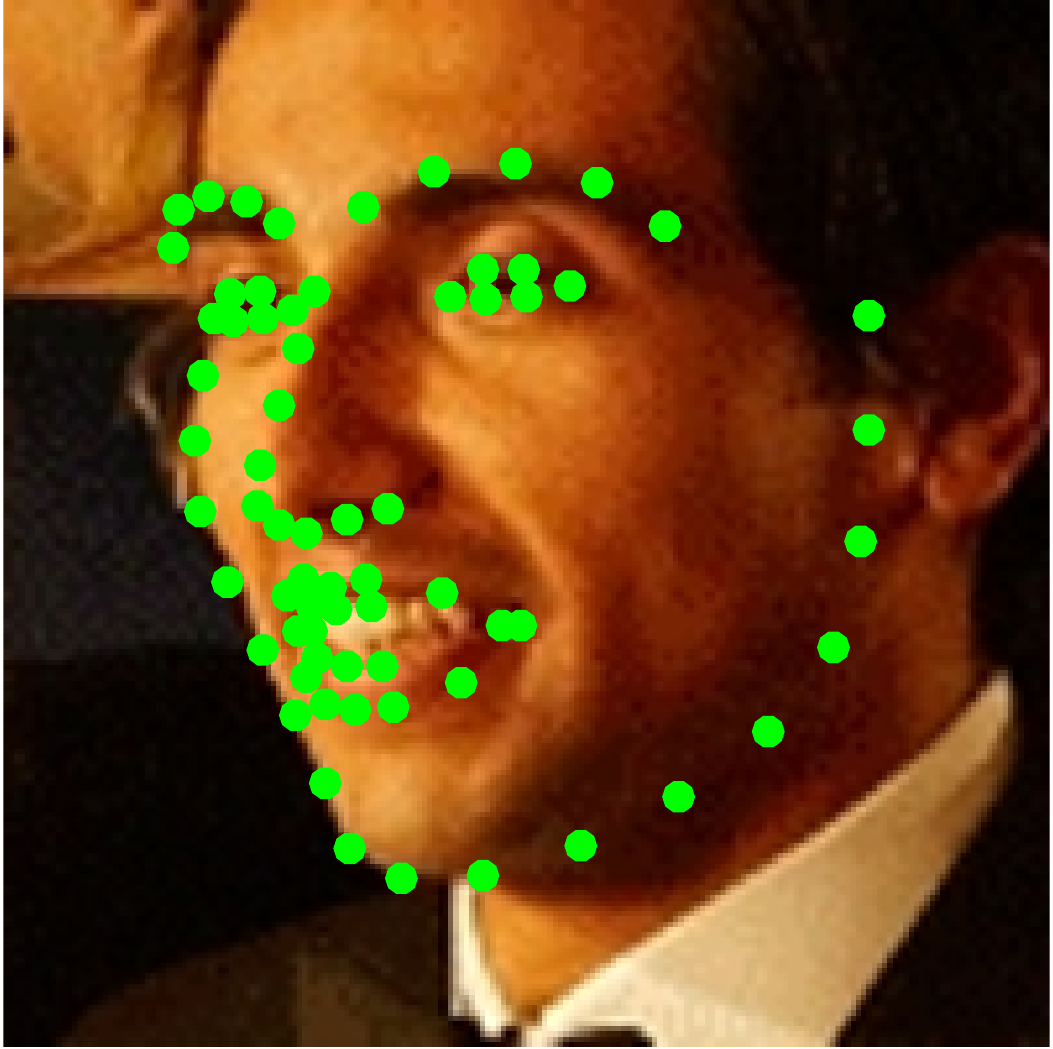}
&
\includegraphics[width=0.82in]{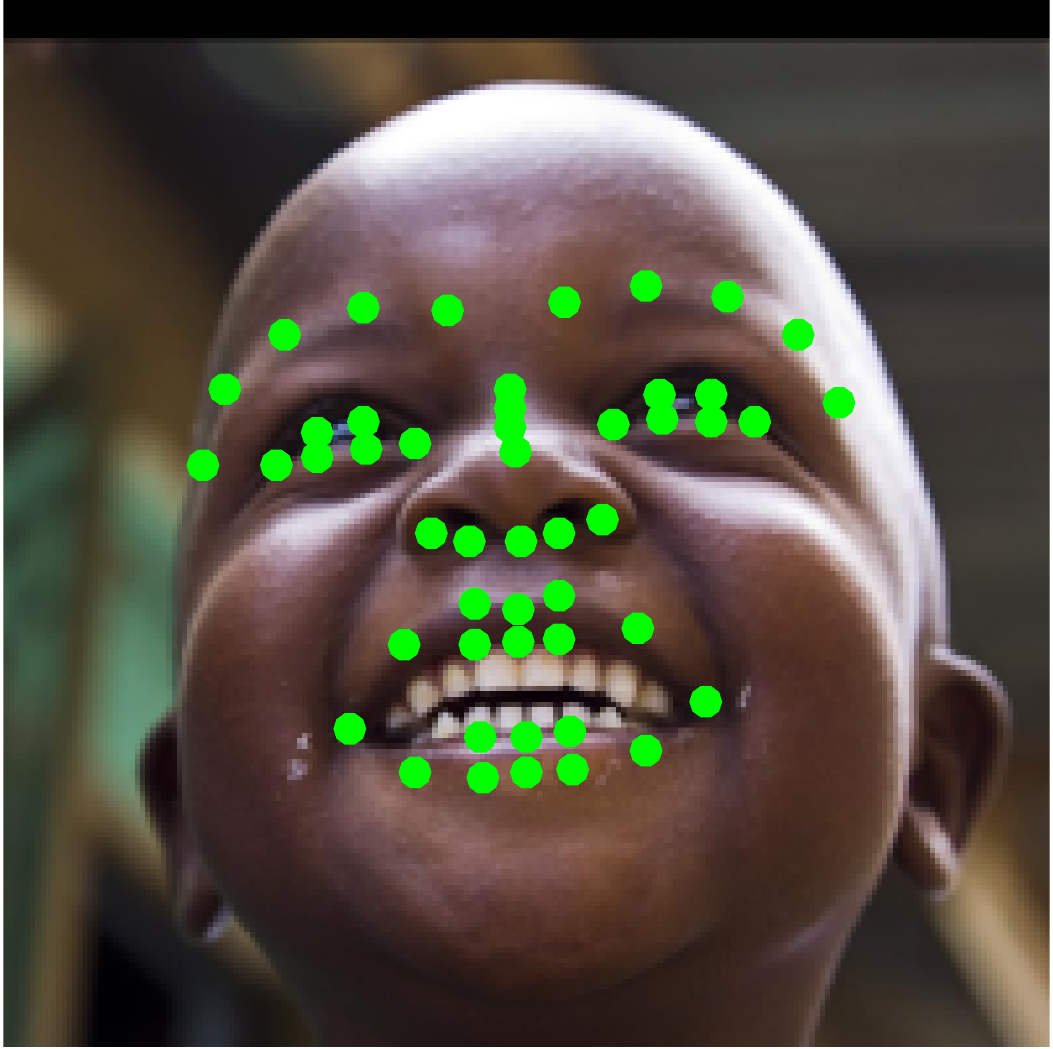}
&
\includegraphics[width=0.82in]{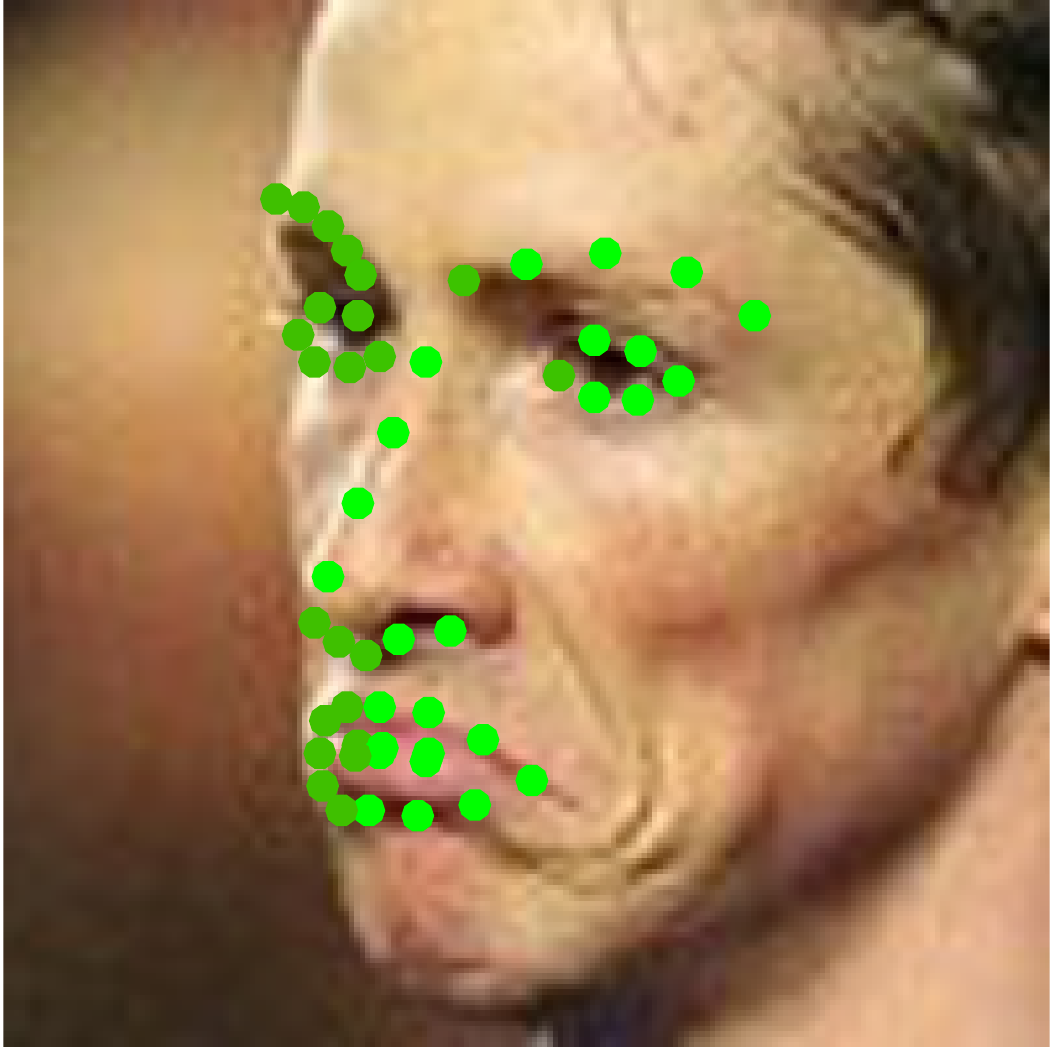}
&
\includegraphics[width=0.82in]{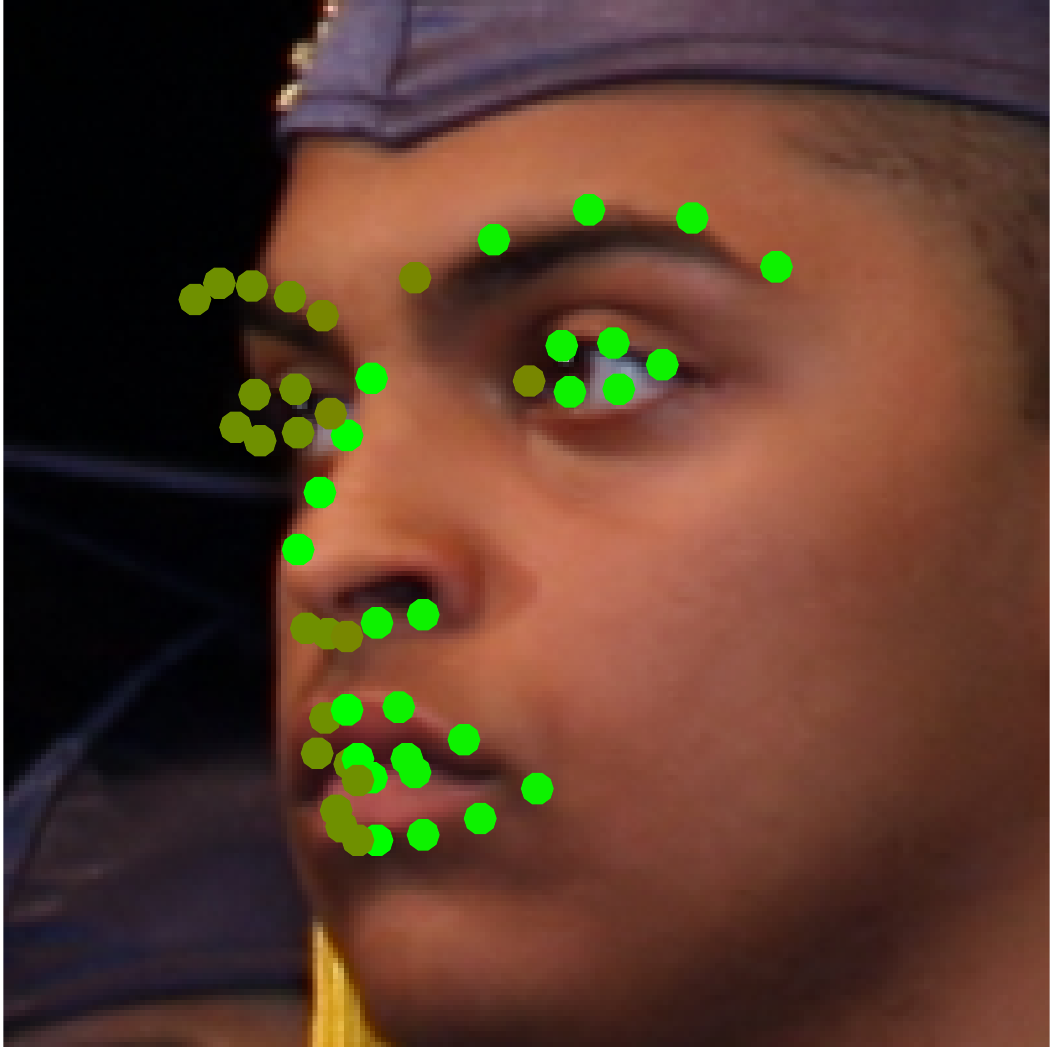}
&
\includegraphics[width=0.82in]{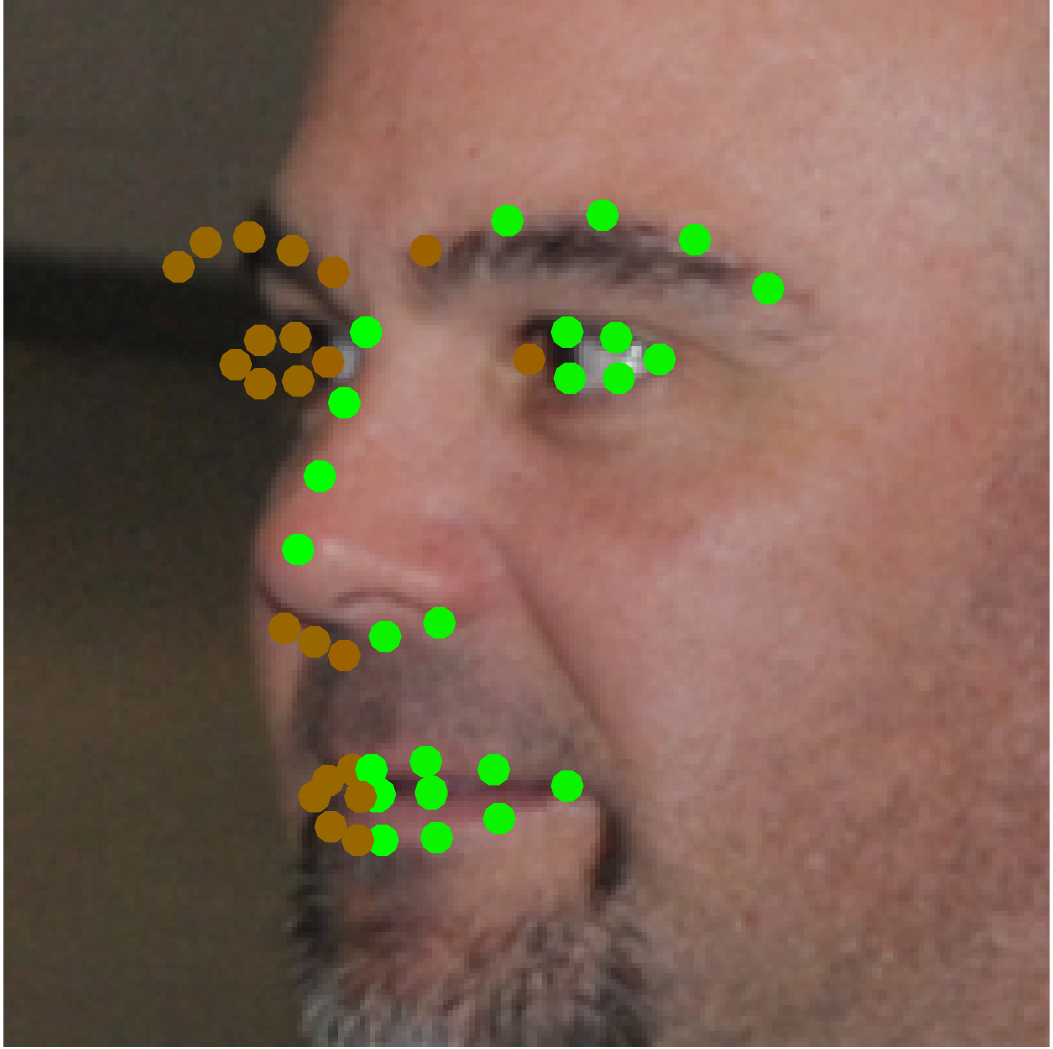}

\\

\includegraphics[width=0.82in]{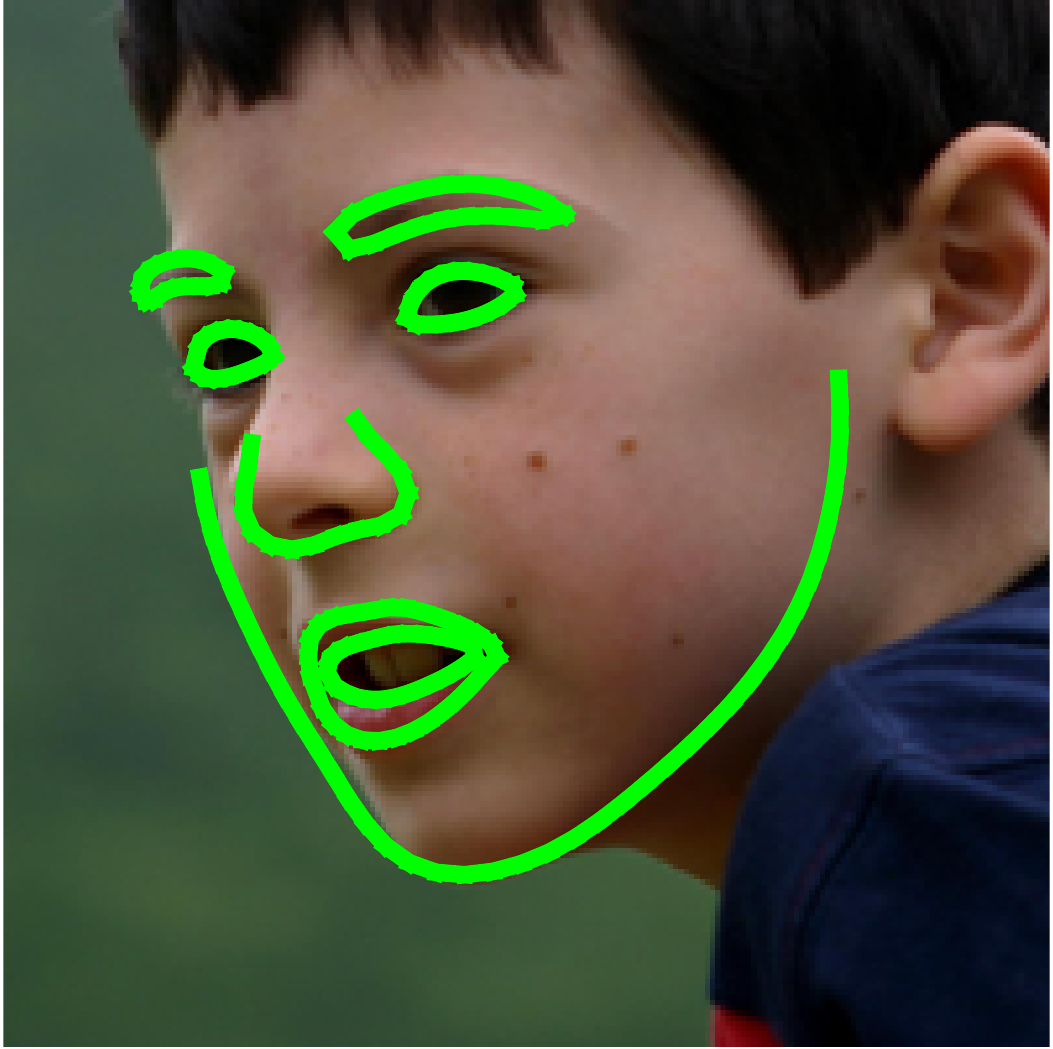}
&
\includegraphics[width=0.82in]{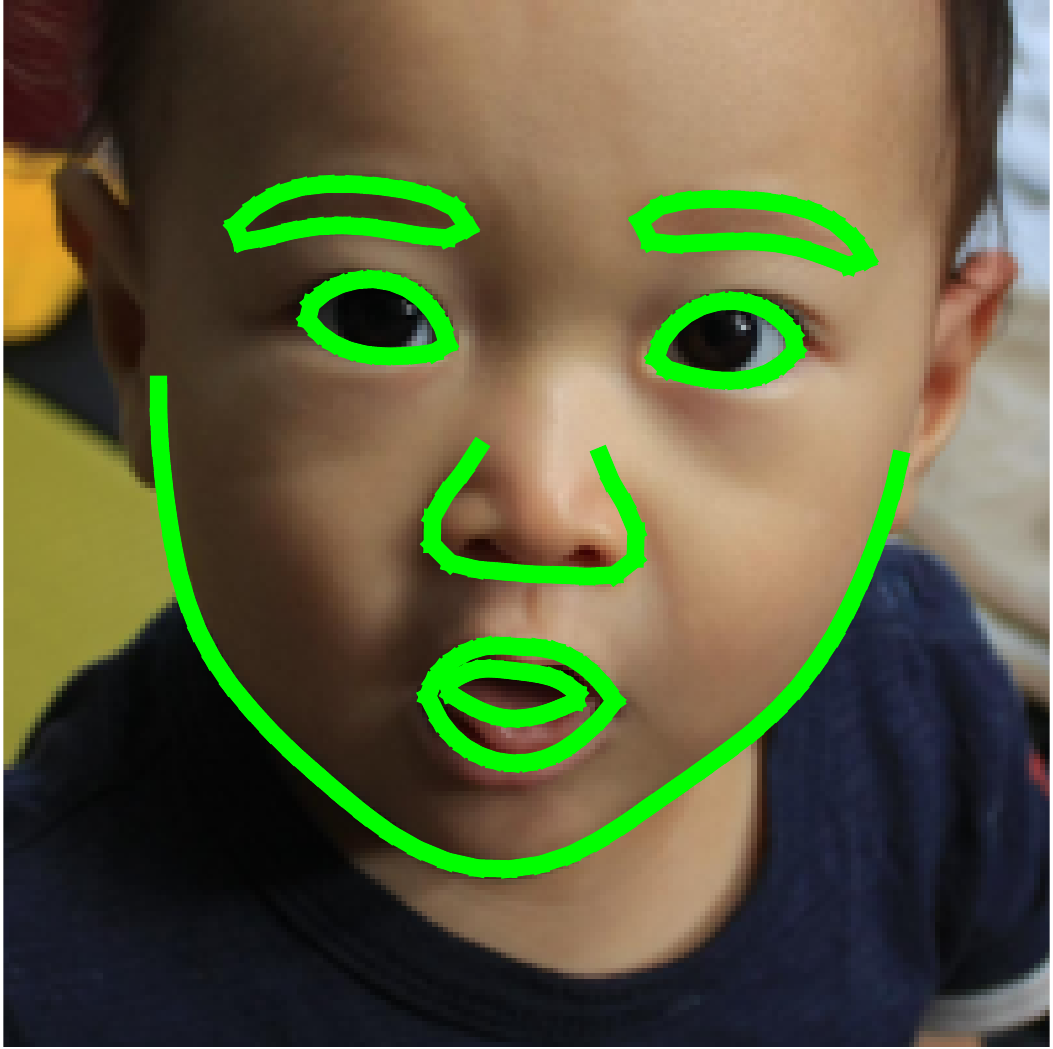}
&
\includegraphics[width=0.82in]{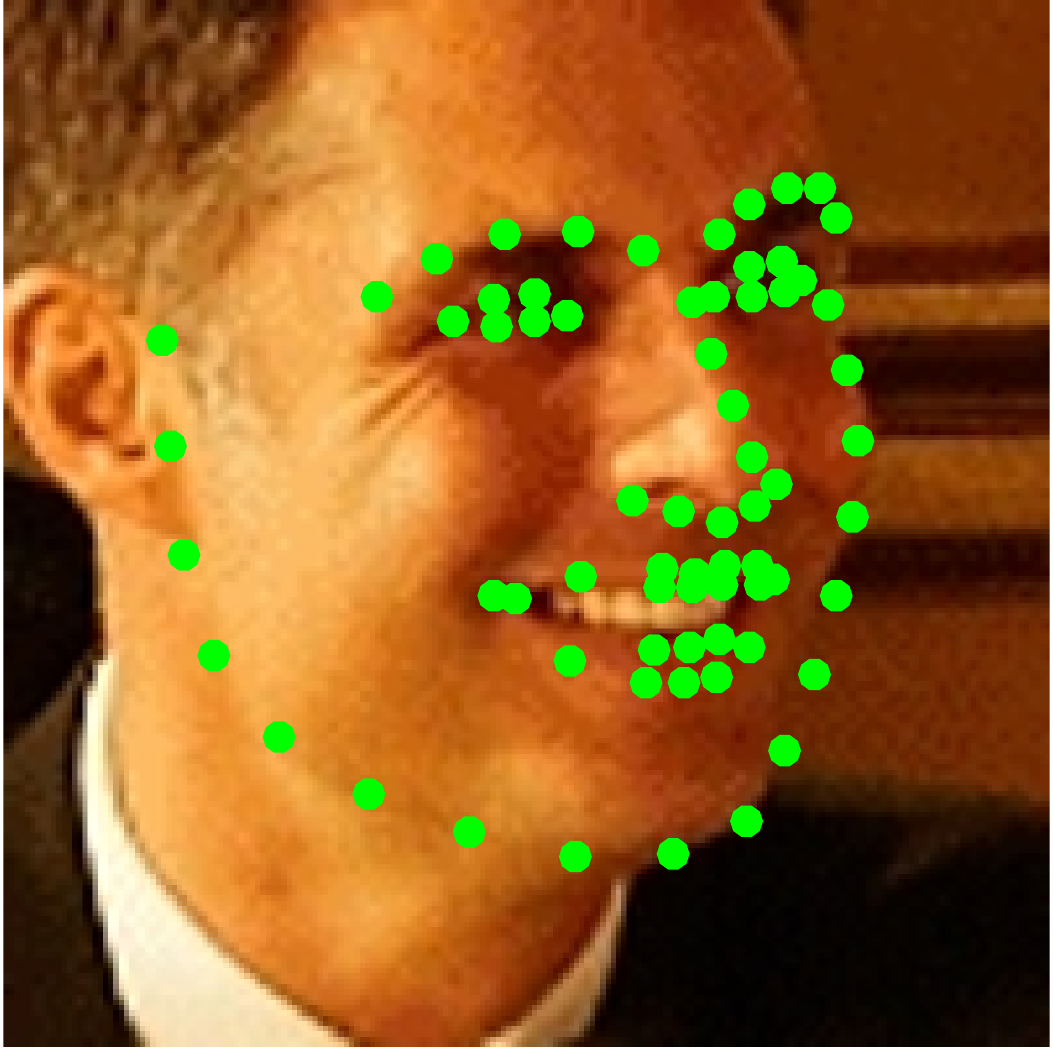}
&
\includegraphics[width=0.82in]{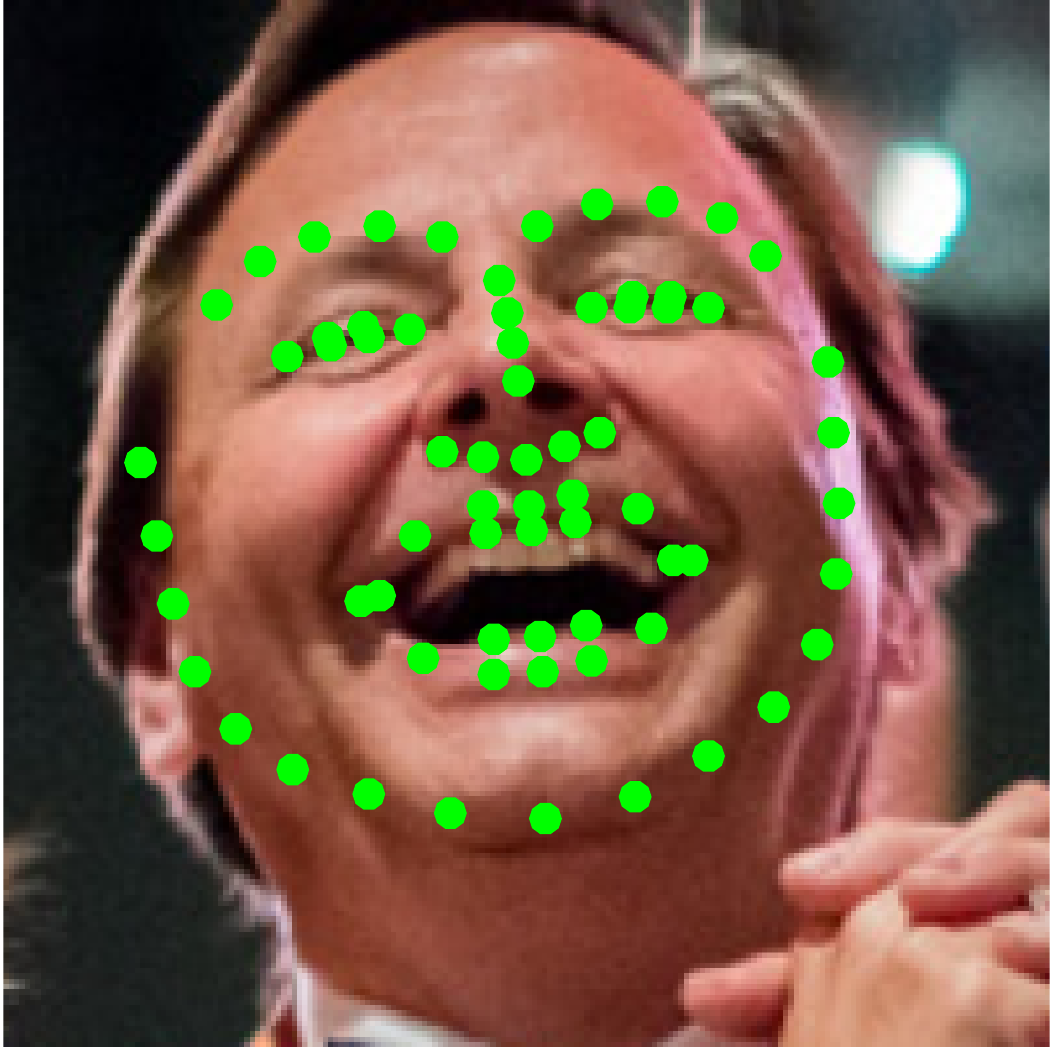}
&
\includegraphics[width=0.82in]{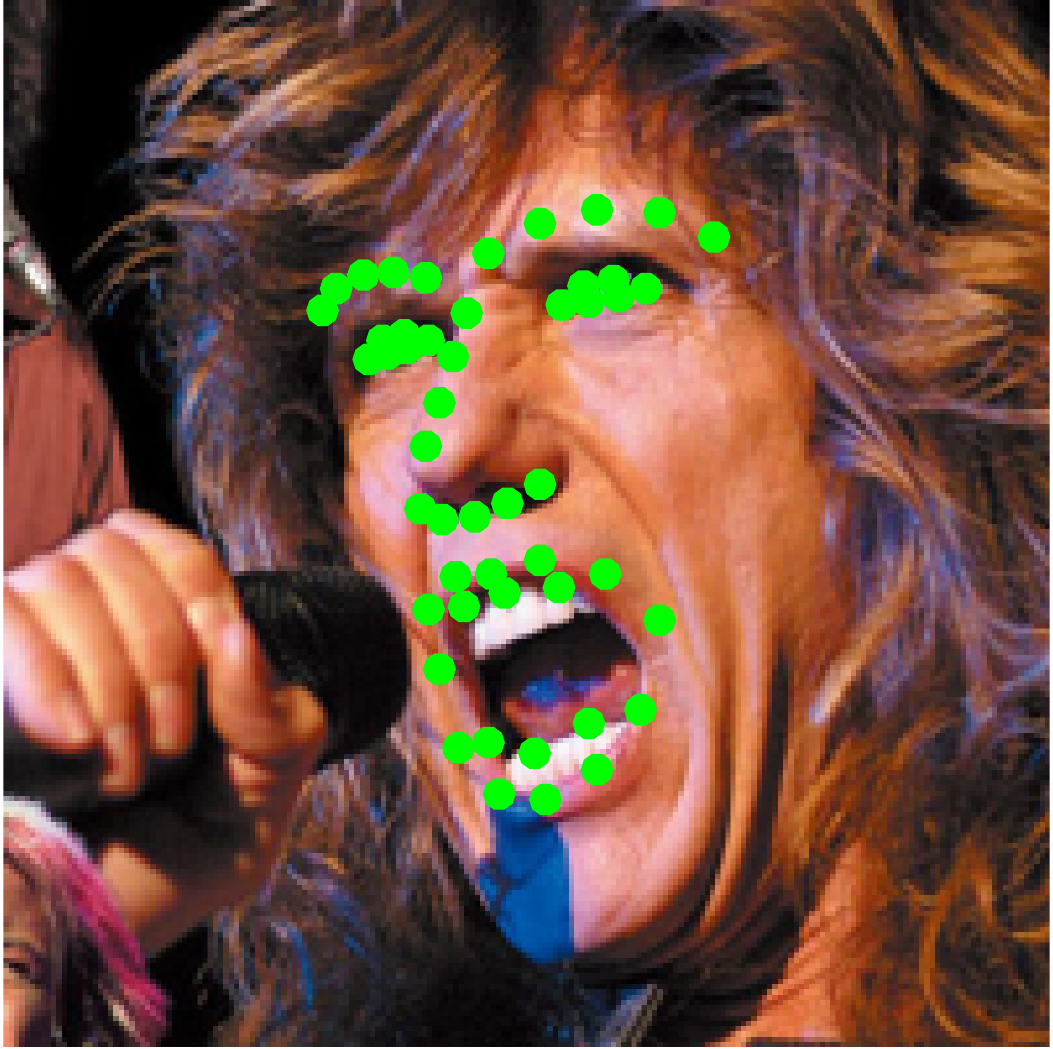}
&
\includegraphics[width=0.82in]{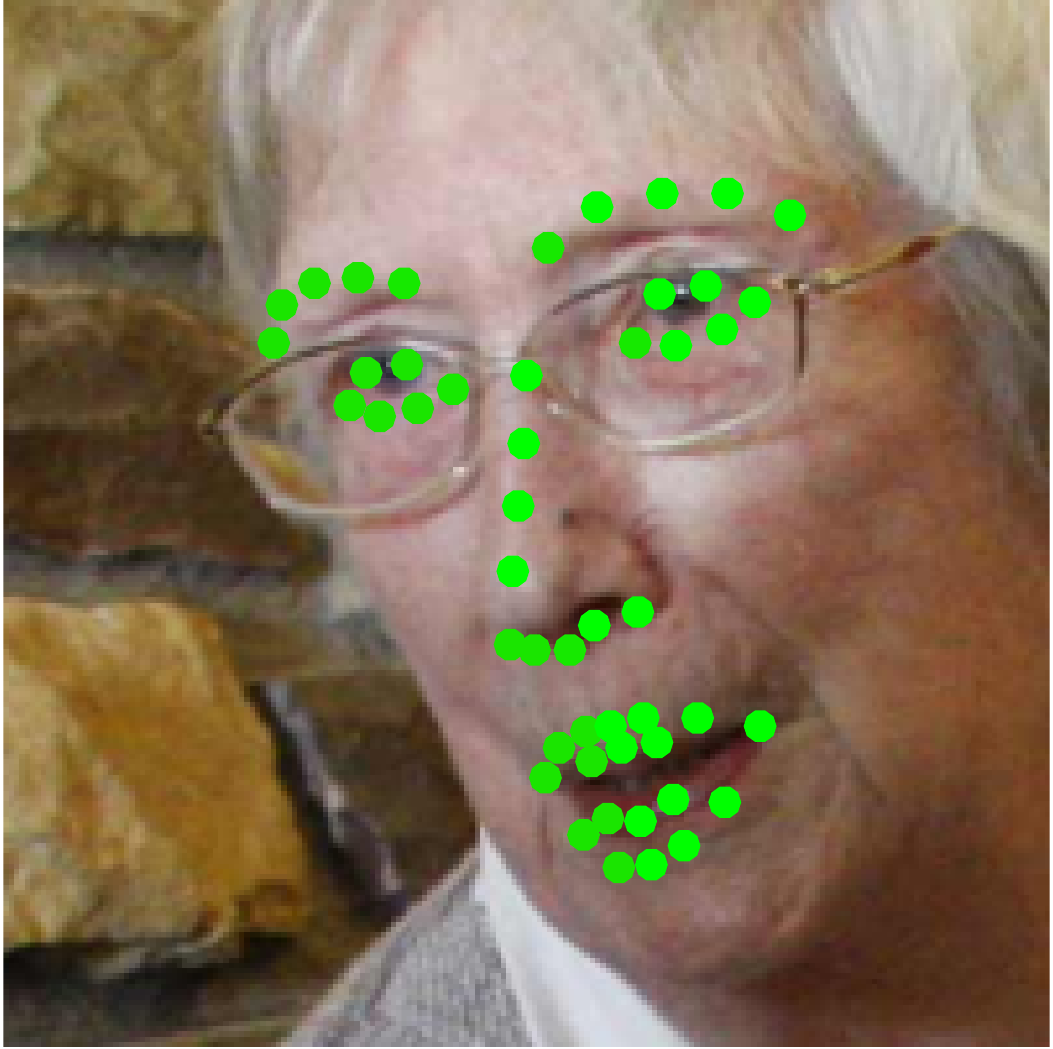}
&
\includegraphics[width=0.82in]{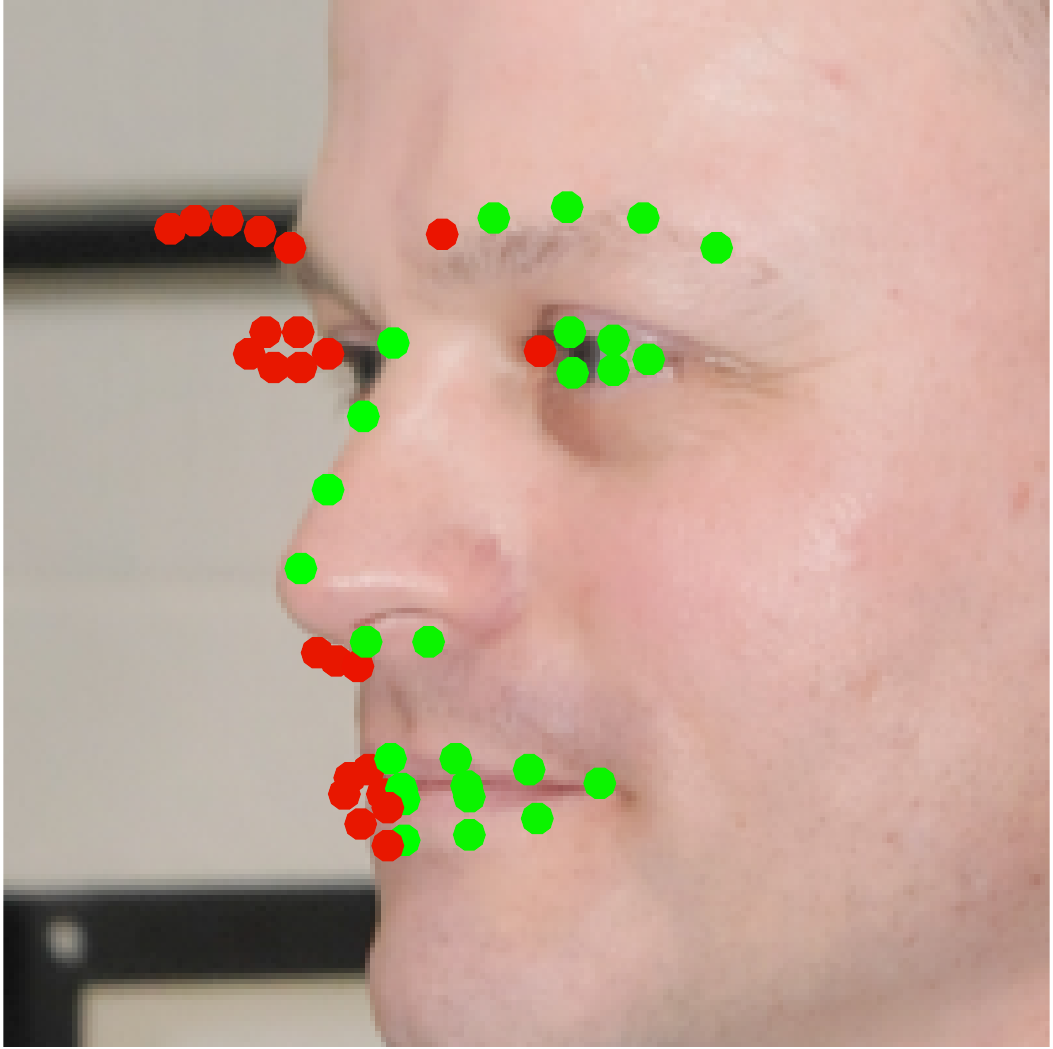}
&
\includegraphics[width=0.82in]{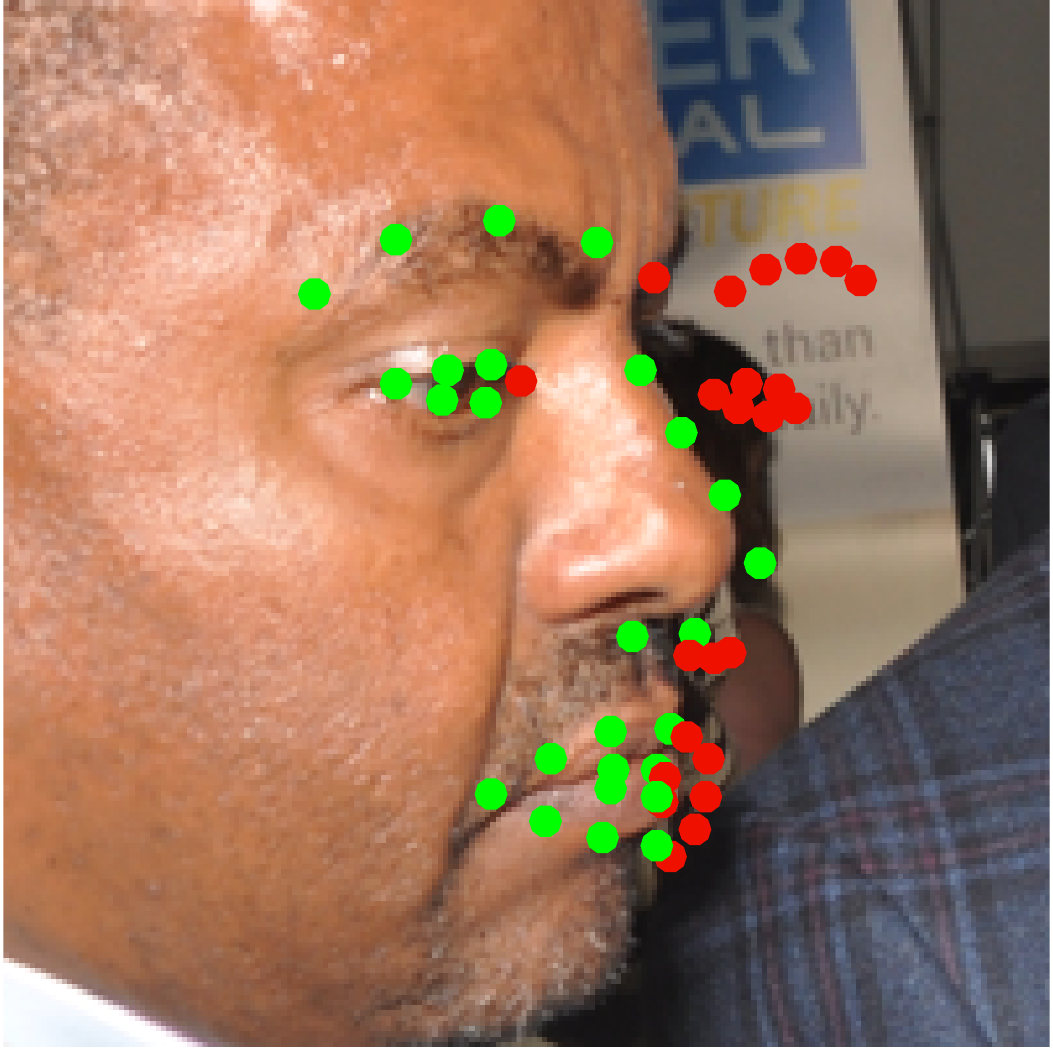}
\\
\includegraphics[width=0.82in]{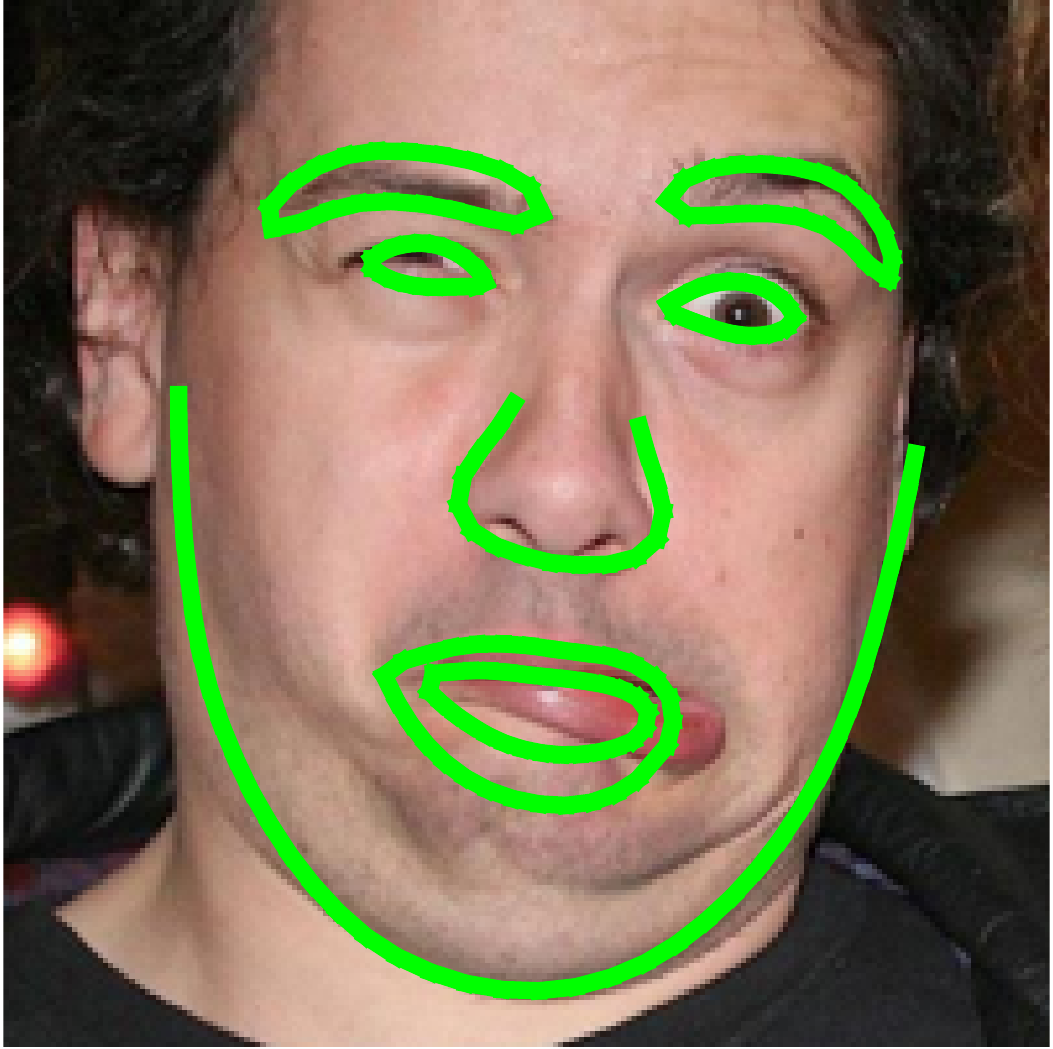}
&
\includegraphics[width=0.82in]{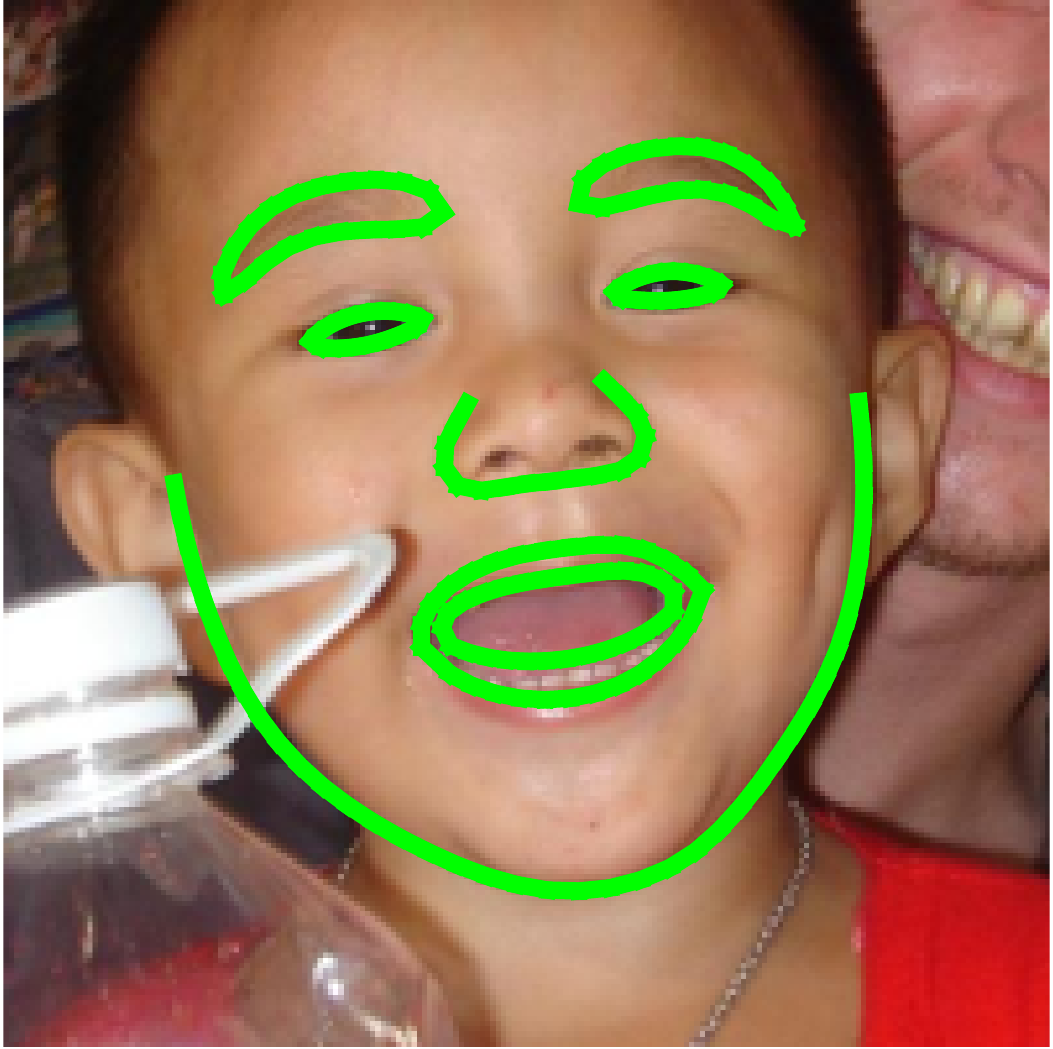}
&
\includegraphics[width=0.82in]{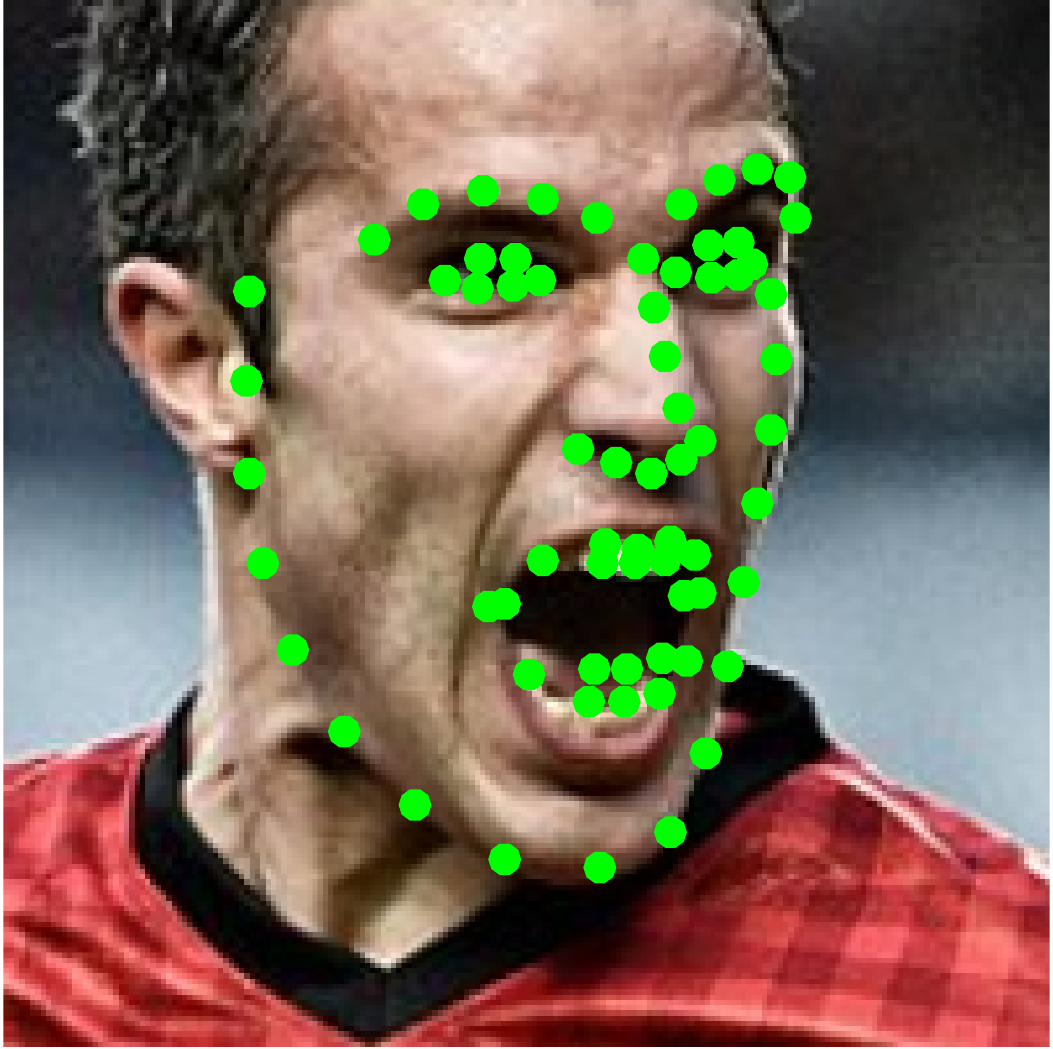}
&
\includegraphics[width=0.82in]{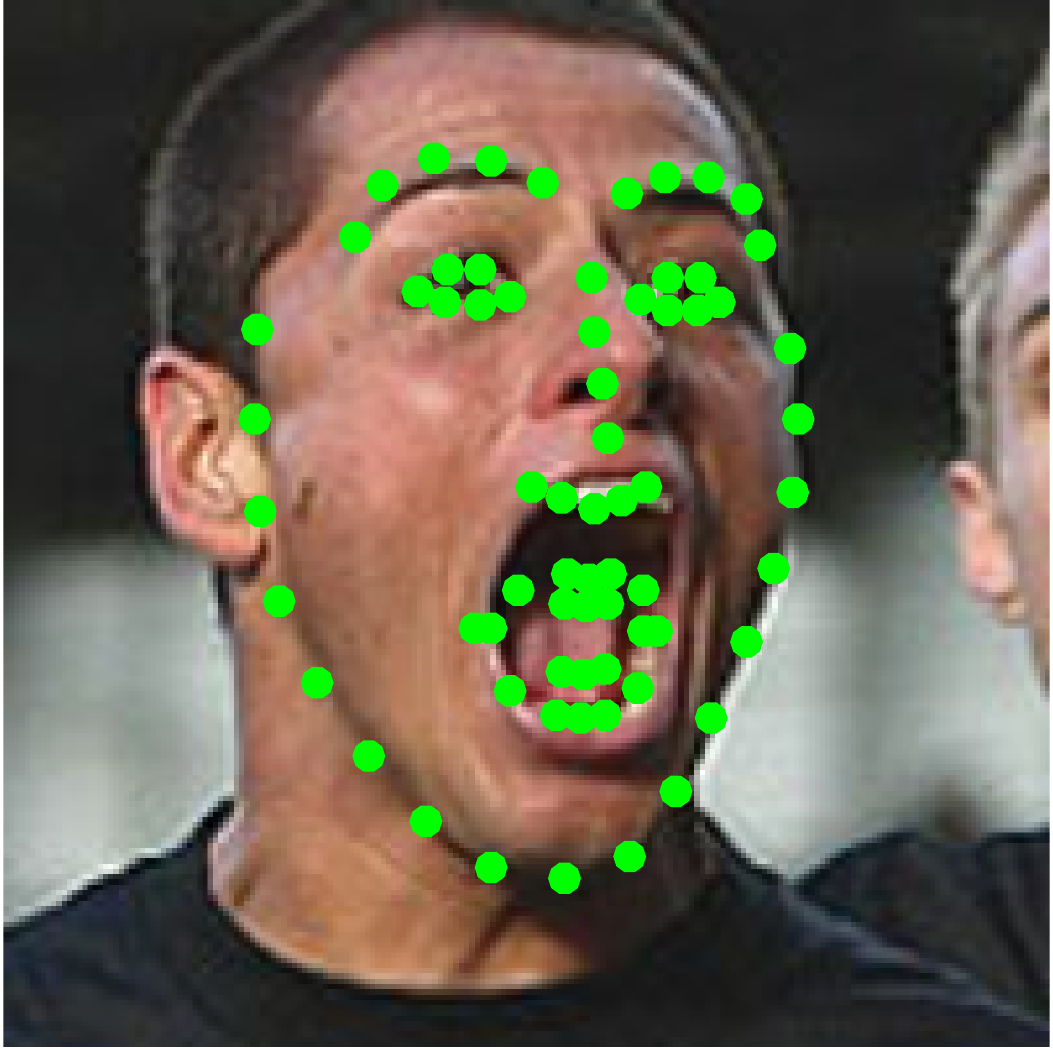}
&
\includegraphics[width=0.82in]{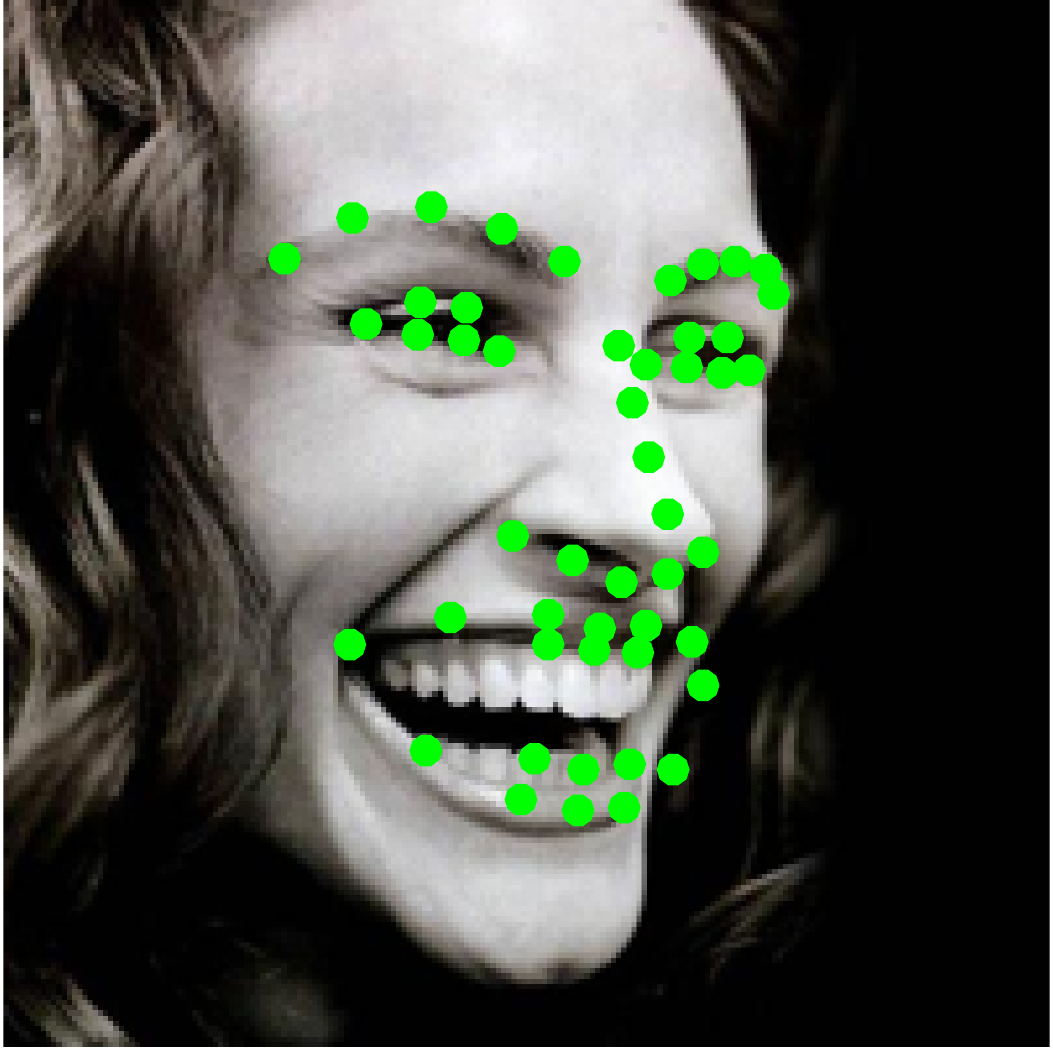}
&
\includegraphics[width=0.82in]{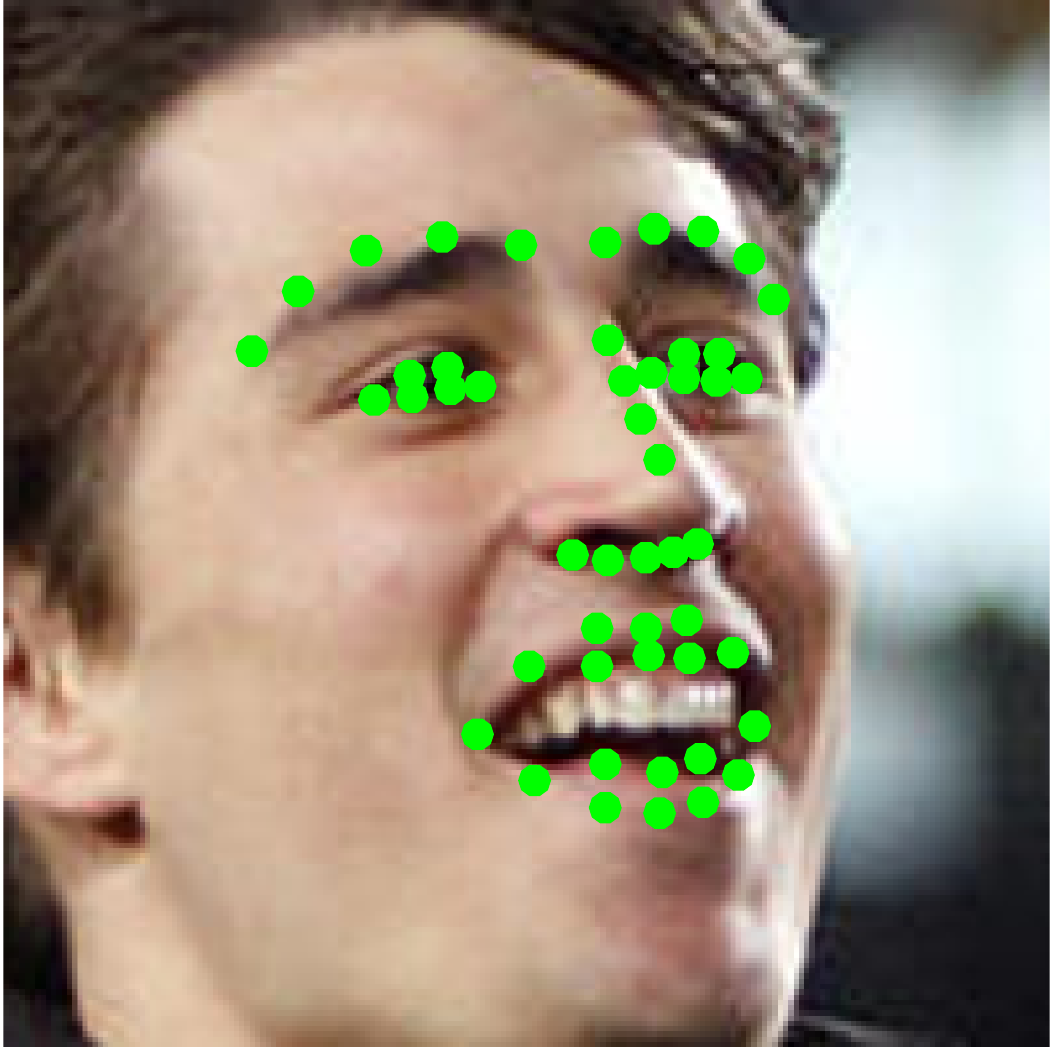}
&
\includegraphics[width=0.82in]{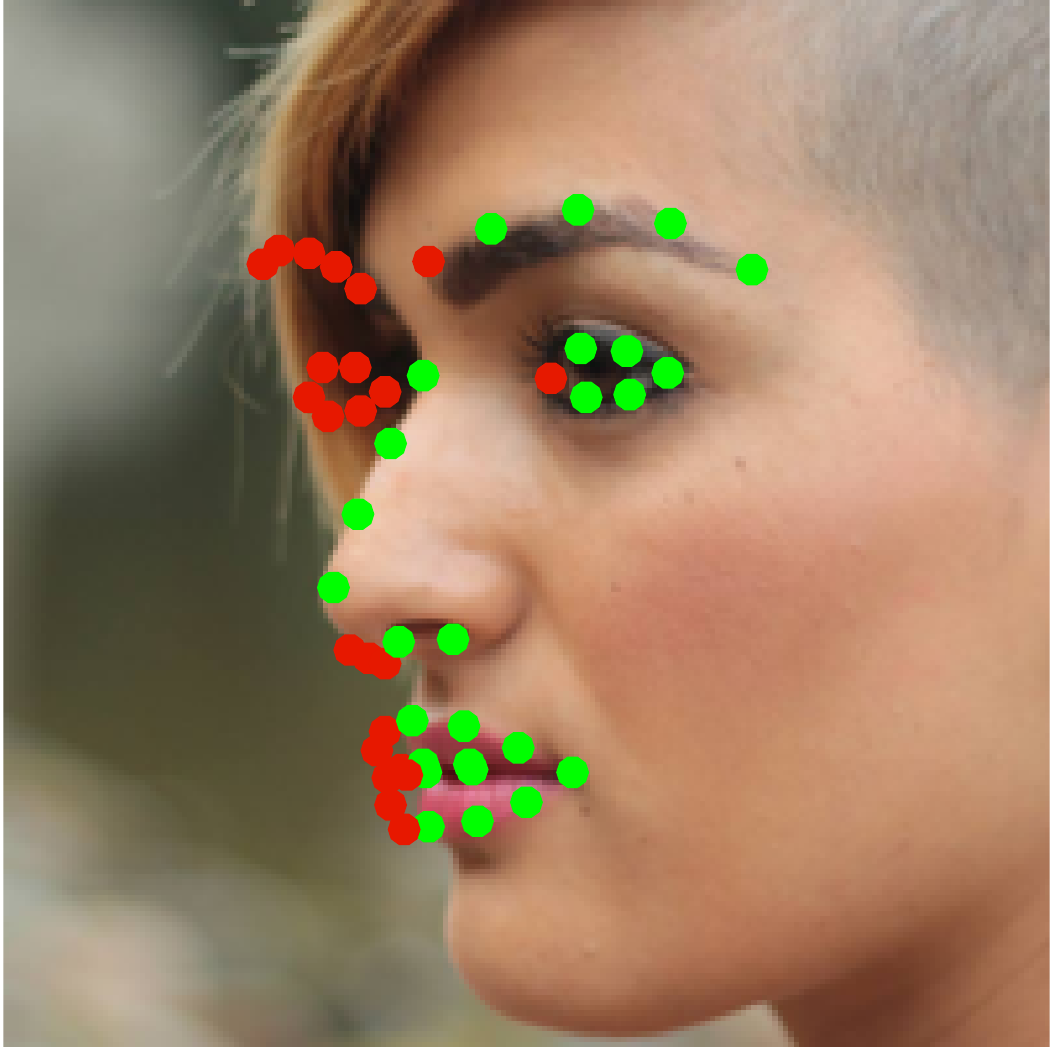}
&
\includegraphics[width=0.82in]{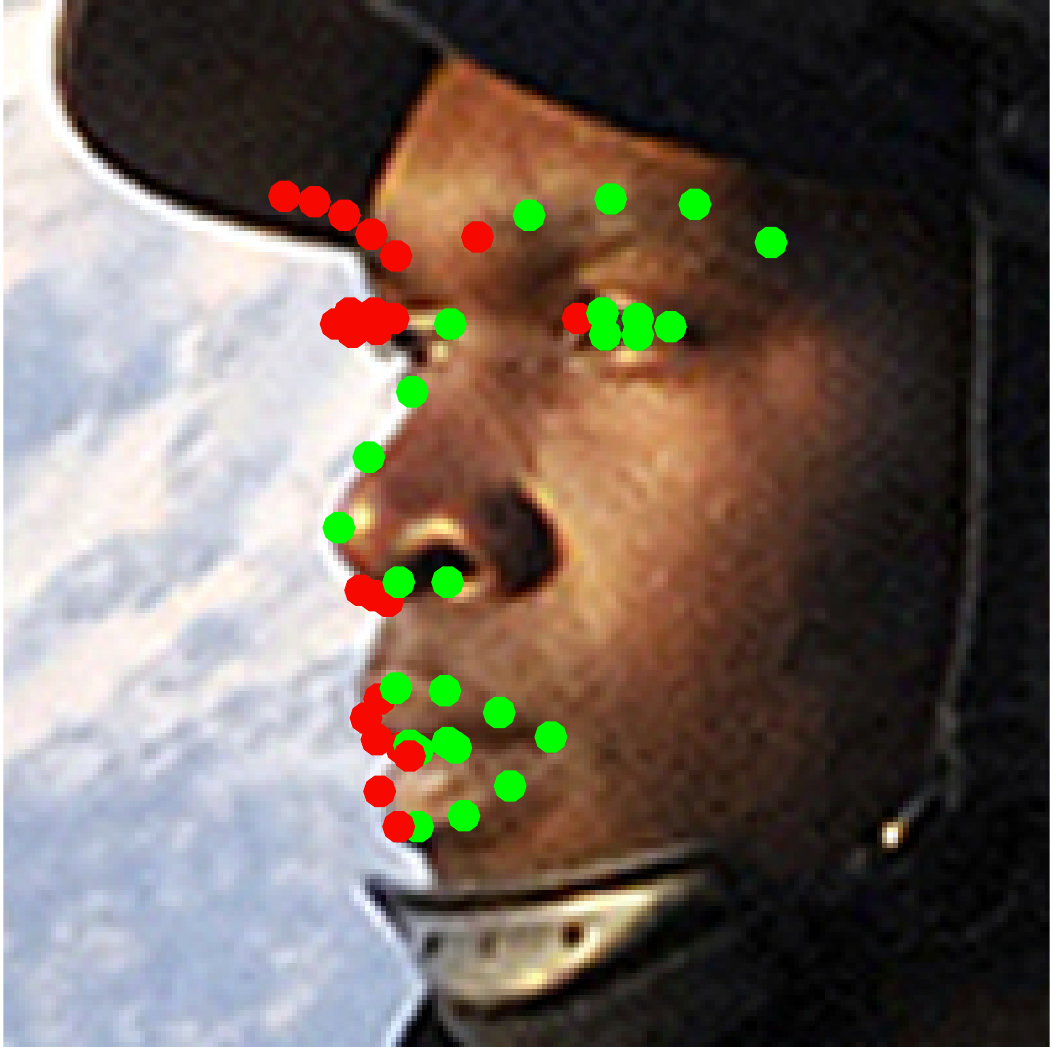}
\end{tabular}
\end{small}
\end{center}
\vspace{-8pt}
\caption{
Selected qualitative results on Helen (columns 1 \& 2), IBUG (columns 3 \& 4), 300-W Fullset (columns 5 \& 6), and our new Flickr-Profile dataset (columns 7 \& 8). 
Columns 5 - 8 were generated using our full model and include pose-related landmark visibility estimation. 
Notice that the red landmarks are correctly labeled as occluded in columns 7 \& 8. 
Please see our supplementary material for additional results.
\textbf{Best viewed electronically in color.}
} \label{fig-qualitative_results} \vspace{8pt}
\end{figure*}

\section{Results and Discussion} \label{results}

\subsection{Datasets} 

\textbf{Multi-PIE} \cite{Gross_IVC2010_Multipie} is an extensive in-the-lab dataset that includes annotations on profile faces. 
We use it to train our most general model, which can handle both frontal and profile faces. 
Additional annotations on Multi-PIE profile faces were provided by \cite{Zhu2012}. 
One limitation of Multi-PIE is that all profile landmark annotations are on neutral-expression faces. 

\textbf{Helen} (194 landmarks) \cite{Le2012} is a dataset of 2330 high-resolution in-the-wild face images. 
Most faces are near-frontal, but other challenges are significant, including exaggerated facial expressions, and diversity of subjects. 

\textbf{300 Faces in-the-wild Challenge (300-W)} (68 landmarks) \cite{faces300w} is a collection of in-the-wild datasets that were re-annotated with landmarks consistent with Multi-PIE. 
We closely followed the training and testing procedures described in previous work \cite{Zhu2015} and used the prescribed face detector results provided by 300-W for initialization. 
``Common Subset'' includes LFPW \cite{Belhumeur2011} and Helen, ``Challenging Subset'' is the iBUG subset, and ``Fullset'' is the union of the three.
We note that ``inter-ocular distance'' is defined by 300-W as the distance between the outer eye corners.

\textbf{FERET} \cite{Phillips1997} is an older dataset that includes neutral-expression profile faces against uniform backgrounds.
Wu and Ji \cite{Wu2015} recently added landmark annotations to $465$ of the profile faces, which allows us to quantitatively compare our results with theirs. 

\textbf{Flickr-Profile} is a new in-the-wild dataset that we assembled to validate our approach. 
Specifically, we collected public domain images from Flickr depicting 170 unique profile faces. 
We annotated the visible portion of the non-jawline landmarks according to the Multi-PIE mark-up. 
Images in \textit{Flickr-Profile} include a variety of challenging illumination conditions, resolutions, noise levels, \etc, as shown in \fig{fig-qualitative_results}. 
%Almost all images depict faces with neutral expressions in order 
%We limited the amount of facial expressions in Flickr-Profile to better reflect the Multi-PIE annotations that we have access to. 
%That is, the public version of Multi-PIE does not include annotations on non-neutral-expression profile faces, and so we cannot adequately train our algorithm to deal with them.

\subsection{Comparison on Popular in-the-wild Datasets}

Few recent methods handle both frontal and profile faces with the same model, which limits our ability to evaluate the most general version of our algorithm with recent work. 
Therefore, we first demonstrate that our method is competitive in a more conventional setting (\eg, with no pose-related landmark occlusion). 
Specifically, we report averaged errors on Helen and 300-W datasets in \tabfig{fig-quantitative_helen_300w}.\footnote{Like others \cite{Ren2014,Zhu2015}, we recognize that the alignment accuracy on LFPW \cite{Belhumeur2011} has saturated, and so we do not present separate results on LFPW.} 
Note that, during training, we assume that all landmarks are visible for these datasets. 
Our method performs favorably compared to state-of-the-art methods, achieving the lowest errors in both datasets.

\subsection{Comparison on Images with Large Head Rotation}

\begin{figure*}
\begin{center}
\begin{small}
\begin{tabular}{c@{ }c@{ }c@{ }c@{ }c | @{\hspace{10pt}}c}
\includegraphics[height=0.9in]{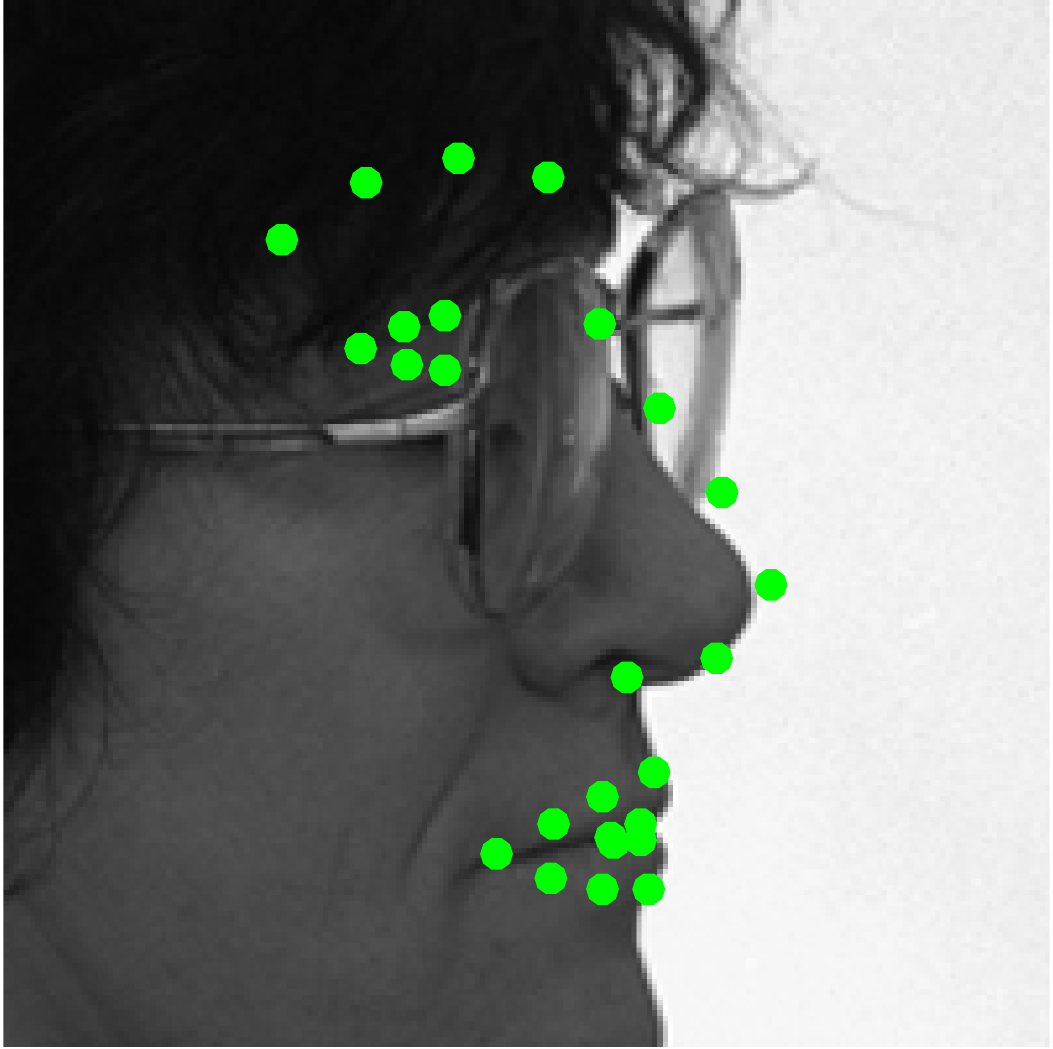}
&
\includegraphics[height=0.9in]{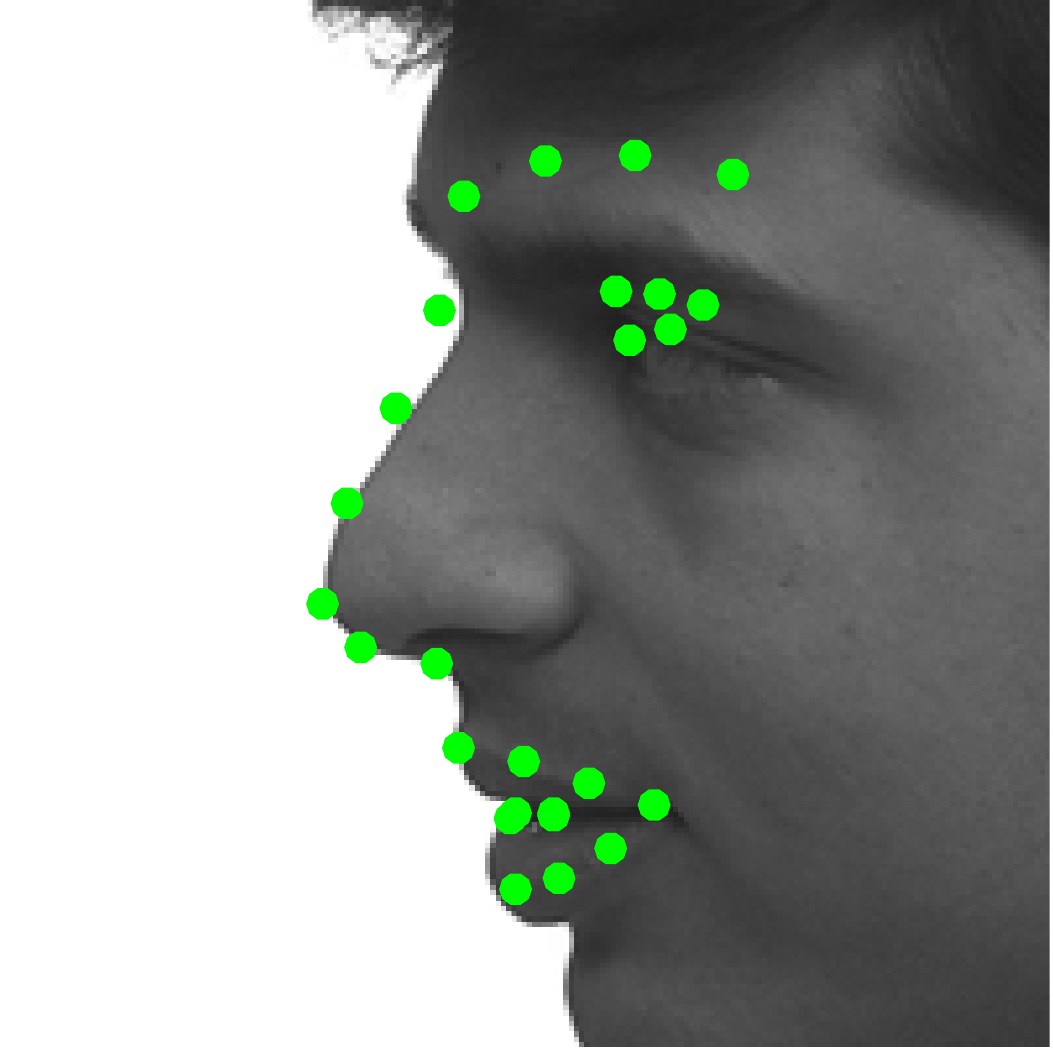}
& 
\includegraphics[height=0.9in]{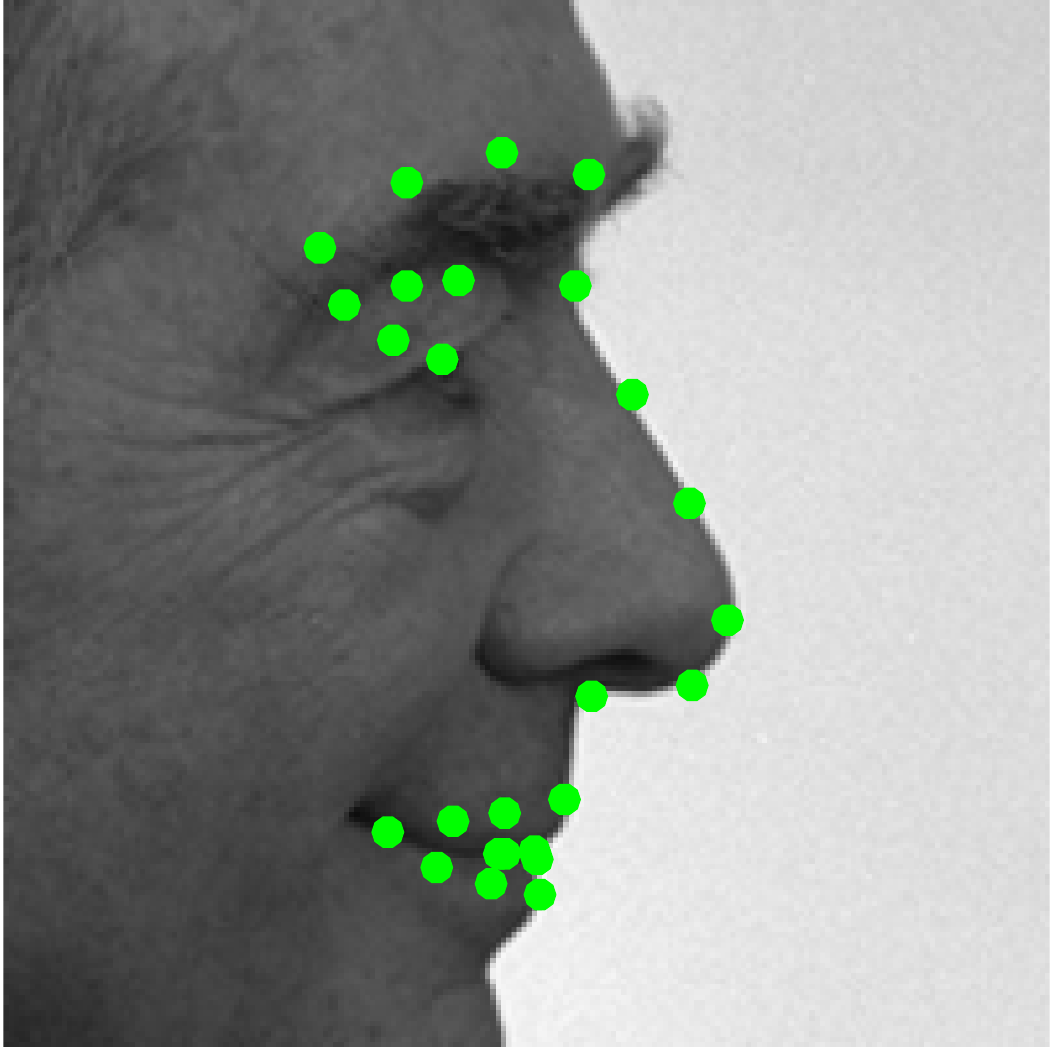}
& 
\includegraphics[height=0.9in]{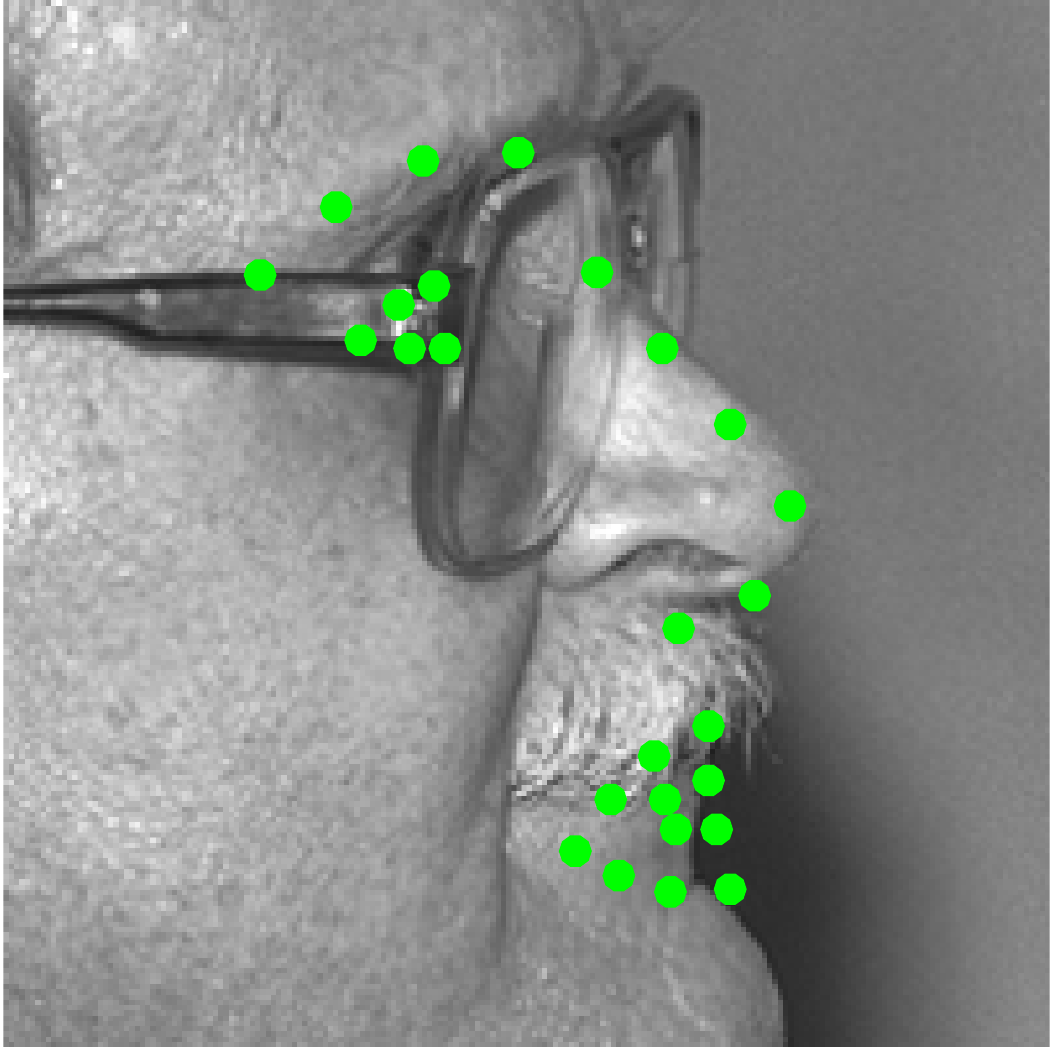}
& 
\includegraphics[height=0.9in]{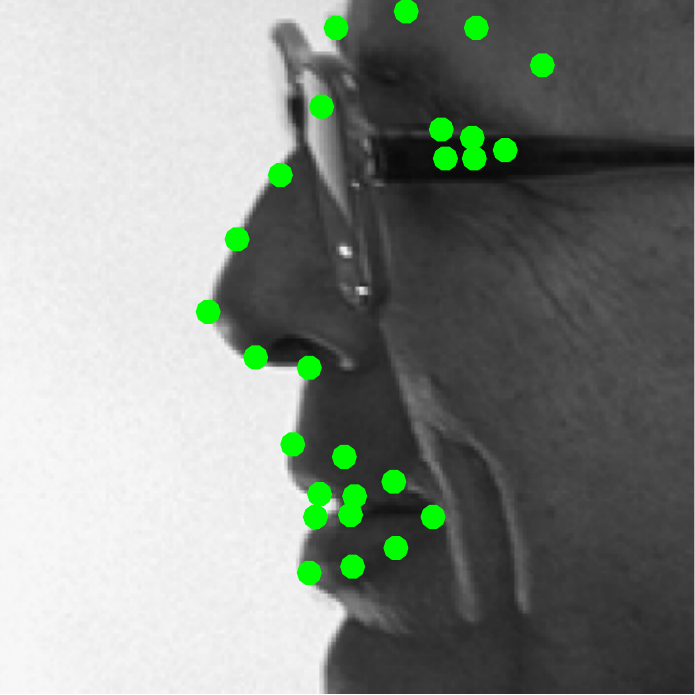}
&
\includegraphics[height=0.9in]{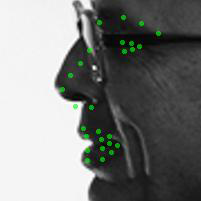}
\\
\multicolumn{5}{c}{(a) Our five worst-fitting results on FERET} & (b) Result from \cite{Wu2015}\\
\end{tabular} \\ 
\vspace{8pt}
\begin{tabular}{r@{\hspace{4pt}}c@{\hspace{12pt}}r@{\hspace{4pt}}c}
%& Left profile & & Right profile \\
\begin{sideways}\hspace{30pt}Fraction of test faces\end{sideways}
&
\includegraphics[width=2.7in]{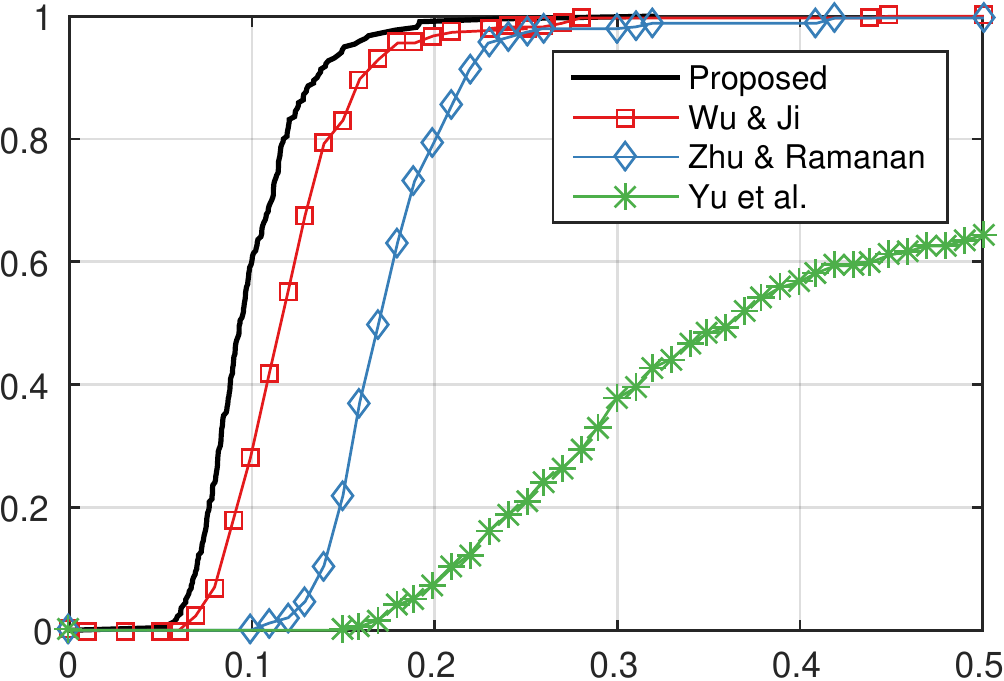}
&
\begin{sideways}\hspace{30pt}Fraction of test faces\end{sideways}
&
\includegraphics[width=2.7in]{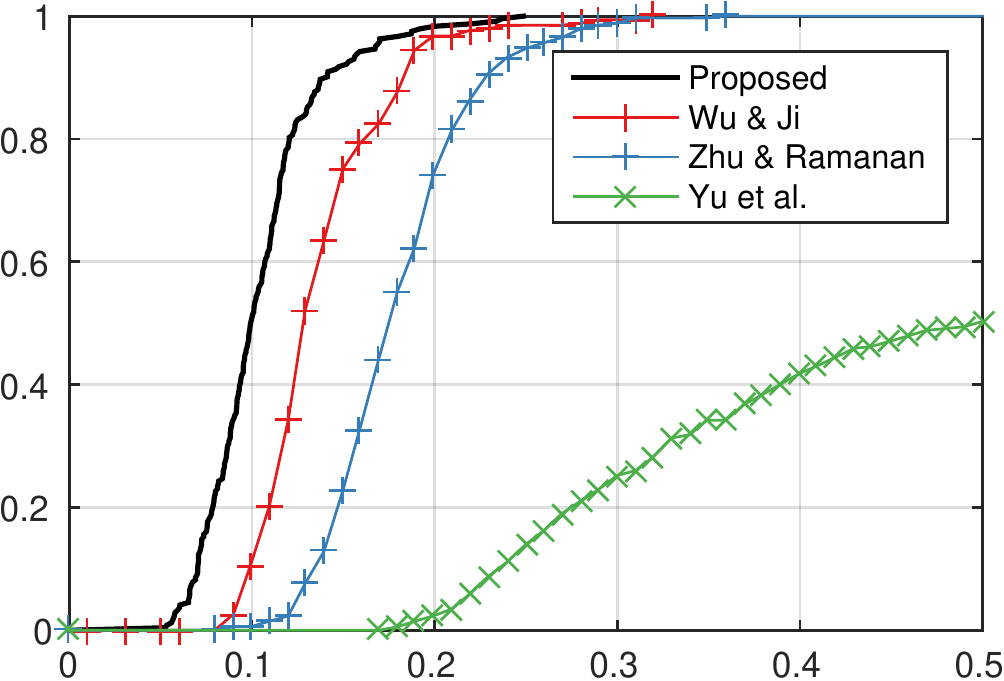} 
\\
& Average normalized error & & Average normalized error \vspace{4pt} \\
& (c) Left profile & & (d) Right profile
\end{tabular}
\end{small}
\end{center}
%\vspace{-8pt}
\caption{
Results on profile faces from the FERET database \cite{Phillips1997}.
(a) The five \textit{worst-fitting} results from our proposed algorithm. 
(b) The single worse-fitting result as reported by Wu and Ji \cite{Wu2015}. 
(c)-(d) Cumulative error distribution (CED) curves on profile faces. 
The CED curves of Wu \& Ji \cite{Wu2015}, Zhu \& Ramanan \cite{Zhu2012}, and Yu \etal \cite{Yu2013} are reproduced here directly from \cite{Wu2015} for comparison. 
We see that the results from our proposed algorithm are significantly more accurate than recent state-of-the-art methods. 
} 
\label{fig-ours_vs_wuiccv2015} %\vspace{-8pt}
\end{figure*}

In \fig{fig-ours_vs_wuiccv2015} we compare our results on FERET with \cite{Wu2015}, \cite{Zhu2012}, and \cite{Yu2013}. 
The curves for \cite{Wu2015}, \cite{Zhu2012}, and \cite{Yu2013} were taken directly from \cite{Wu2015}. 
Like \cite{Wu2015}, we use Multi-PIE, LFPW, and Helen to train our model for testing on FERET, and we similarly use half the distance between the eye and mouth corner points to normalize each mean error. 
We see that our results compare favorably. 

\begin{figure}
\begin{center}
\begin{small}
\begin{tabular}{r@{\hspace{4pt}}c}
%& Left profile & & Right profile \\
\begin{sideways}\hspace{30pt}Fraction of test faces\end{sideways}
&
\includegraphics[width=2.7in]{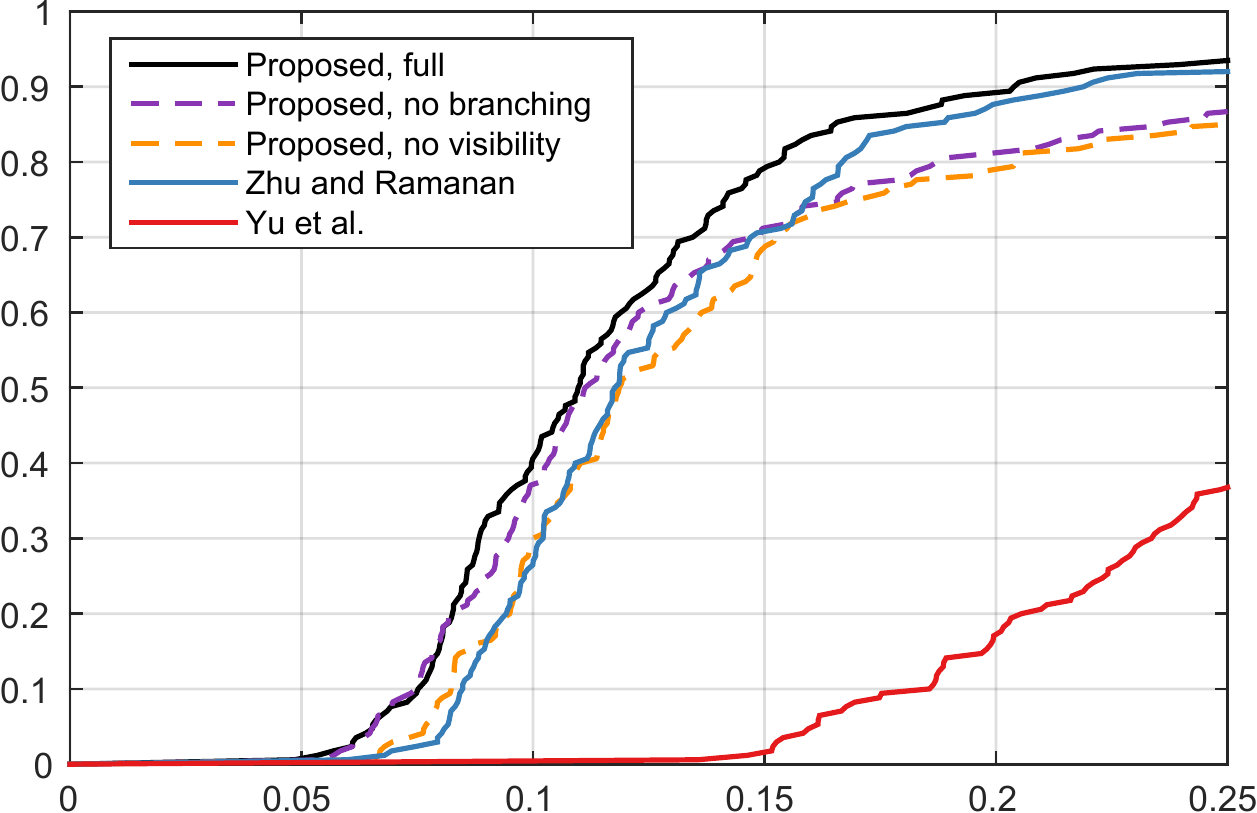}
\\
& Average normalized error \vspace{4pt}
\end{tabular}
\end{small}
\end{center}
\caption{
Results on Flickr-Profile faces. 
We see that our proposed algorithm produces more accurate results when branching and landmark visibility estimation are included. 
} 
\label{fig-flickr_ours_vs_others} %\vspace{-8pt}
\end{figure}

In \fig{fig-flickr_ours_vs_others} we validate our algorithm on Flickr-Profile and compare with two other baseline algorithms. 
We ran the executables from \cite{Zhu2012} and \cite{Yu2013} to generate the plots in \fig{fig-flickr_ours_vs_others}.
Our full algorithm performs well compared to both. 
To initialize \cite{Yu2013}, we scaled, translated, and rotated a mean face shape on top of the ground truth points. 
We ran the code from \cite{Zhu2012} using their slowest (seconds to minutes per frame), but most accurate model. 
In \fig{fig-flickr_ours_vs_others} we also compare the accuracy of our full model with our model without landmark visibility estimation, and our model without branching. 
We see that both are required to achieve the best performance on profile faces using general models.

\subsection{Implementation Details and Runtime}

For all experiments, we use $600$ regression trees with a maximum depth of $6$ for our feature mapping functions, and all of our BCRs have $4$ cascade levels. 
Like previous work, we augment the training datasets by sampling multiple initializations for each image. 
Like \cite{Kazemi2014,Ren2014}, our algorithm is very fast, although our experimental MATLAB implementation is not. 
It runs at about $15$ FPS. 
%No steps in our pipeline consitute significant increases in computation above \cite{Kazemi2014,Ren2014}. 

\begin{table*}
\begin{center}
\begin{small}
%\begin{tabular}{|@{\hspace{3pt}}c@{\hspace{3pt}}|@{\hspace{3pt}}c@{\hspace{3pt}}|@{\hspace{3pt}}c@{\hspace{3pt}}|@{\hspace{3pt}}c@{\hspace{3pt}}|@{\hspace{3pt}}c@{\hspace{3pt}}|@{\hspace{3pt}}c@{\hspace{3pt}}|}
\begin{tabular}{l@{\hspace{-8pt}}c@{\hspace{-4pt}}c@{\hspace{6pt}}c@{\hspace{6pt}}c}
 & Helen & \multicolumn{3}{c}{300-W Dataset} \\
 & (194 landmarks) & \multicolumn{3}{c}{(68 landmarks)} \\
\hline 
Method & & Common & Challenging & Fullset \\
       & & Subset & Subset & \\
\hline 
Zhu \& Ramanan [40] & - & 8.22 & 18.33 & 10.20\\
DRMF [1] & - & 6.65 & 19.79 & 9.22 \\
ESR [8] & 5.70 & 5.28 & 17.00 & 7.58 \\
RCPR [5] & 6.50 & 6.18 & 17.26 & 8.35 \\
SDM [33] & 5.85 & 5.60 & 15.40 & 7.52 \\
ERT [20] & - & - & - & 6.40 \\
LBF [27] & 5.41 & 4.95 & 11.98 & 6.32 \\
LBF fast [27] & 5.80 & 5.38 & 15.50 & 7.37 \\
CFSS [39] & 4.74 & 4.73 & 9.98 & 5.76 \\
CFSS practical [39] & 4.84 & 4.79 & 10.92 & 5.99 \\
Lee \etal [22] & - & - & - & 5.71 \\
Wu \& Ji [32] & - & - & 11.52 & -\\
\hline
Ours & 5.74 & 5.18 & 13.26 & 6.76 \\
\hline 
\end{tabular}
\end{small}
\end{center}
\vspace{-4pt}
\caption{
Averaged normalized error values on popular near-frontal face datasets as reported in the literature. 
Note that our results are comparable to LBF (on which our approach is based), which is competative with the state of the art. 
In conjunction with Figures \ref{fig-flickr_ours_vs_others} and \ref{fig-ours_vs_wuiccv2015}, these results demonstrate that our approach can handle profile faces without sacrificing accuracy on near-frontal faces.  
} \label{fig-quantitative_helen_300w}
\end{table*}

\section{Conclusion and Future Work}

In this work, we have proposed a novel way to augment cascaded shape regression. 
Specifically, we proposed \textit{branching cascaded regressors} for face alignment, in which the algorithm chooses which sequences of regressors to apply adaptively. 
To deal with pose-related landmark visibility, we incorporated SPDMs into the cascaded regression framework, and showed empirically that nonparametric shape models are no better than parametric ones. 
Using our approach, we demonstrated state-of-the-art accuracy on challenging datasets, including those with full-profile faces. 
We hope our work inspires others to address the problem of face alignment on faces in-the-wild with arbitrary head rotation. 
In particular, additional public datasets are needed for training and testing.

\section{Acknowlegements}
\label{sec:ack}
This work was supported in part by the U.S.~Department of Transportation Federal Highway Administration DTFH61-14-C-00011. 

{\small
\bibliographystyle{ieee}
\bibliography{0153}
}

\end{document}